\documentclass[9pt,twocolumn,twoside]{osajnl}

\usepackage{times}

\usepackage{amsmath}
\usepackage{amssymb}
\usepackage{multirow}
\usepackage{multirow}
\usepackage{mathtools}
\usepackage{units}
\usepackage{subfloat}
\usepackage{float}
\usepackage[]{units}
\usepackage{xcolor}
\usepackage{url}
\usepackage{bbm}
\usepackage{algorithm}
\usepackage{algpseudocode}
\usepackage{pbox}
\usepackage{placeins}

\journal{ol} 

\setboolean{shortarticle}{false}

\title{Learning Quadrupedal Locomotion over Challenging Terrain}

\author[1,*]{Joonho Lee}
\author[1,2,$\dagger$]{Jemin Hwangbo}
\author[1]{Lorenz Wellhausen}
\author[3]{Vladlen Koltun}
\author[1]{Marco Hutter}

\affil[1]{Robotic Systems Lab, ETH Zurich, Zurich, Switzerland}
\affil[2]{Robotics and Artificial Intelligence Lab, KAIST, Daejeon, Korea}
\affil[3]{Intelligent Systems Lab, Intel, Santa Clara, CA, USA}
\affil[$\dagger$]{Substantial part of the work was carried out during his stay at 1}
\affil[*]{Corresponding author: jolee@ethz.ch}

\dates{This is the accepted version of Science Robotics Vol. 5, Issue 47, eabc5986 (2020) \\ DOI: 10.1126/scirobotics.abc5986}
\begin{abstract}
Some of the most challenging environments on our planet are accessible to quadrupedal animals but remain out of reach for autonomous machines. Legged locomotion can dramatically expand the operational domains of robotics. However, conventional controllers for legged locomotion are based on elaborate state machines that explicitly trigger the execution of motion primitives and reflexes. These designs have escalated in complexity while falling short of the generality and robustness of animal locomotion. Here we present a radically robust controller for legged locomotion in challenging natural environments. We present a novel solution to incorporating proprioceptive feedback in locomotion control and demonstrate remarkable zero-shot generalization from simulation to natural environments. The controller is trained by reinforcement learning in simulation. It is based on a neural network that acts on a stream of proprioceptive signals. The trained controller has taken two generations of quadrupedal ANYmal robots to a variety of natural environments that are beyond the reach of prior published work in legged locomotion. The controller retains its robustness under conditions that have never been encountered during training: deformable terrain such as mud and snow, dynamic footholds such as rubble, and overground impediments such as thick vegetation and gushing water. The presented work opens new frontiers for robotics and indicates that radical robustness in natural environments can be achieved by training in much simpler domains.
\end{abstract}

\begin{document}

\maketitle
\begin{figure*}
    \centering
    \includegraphics[width=\textwidth]{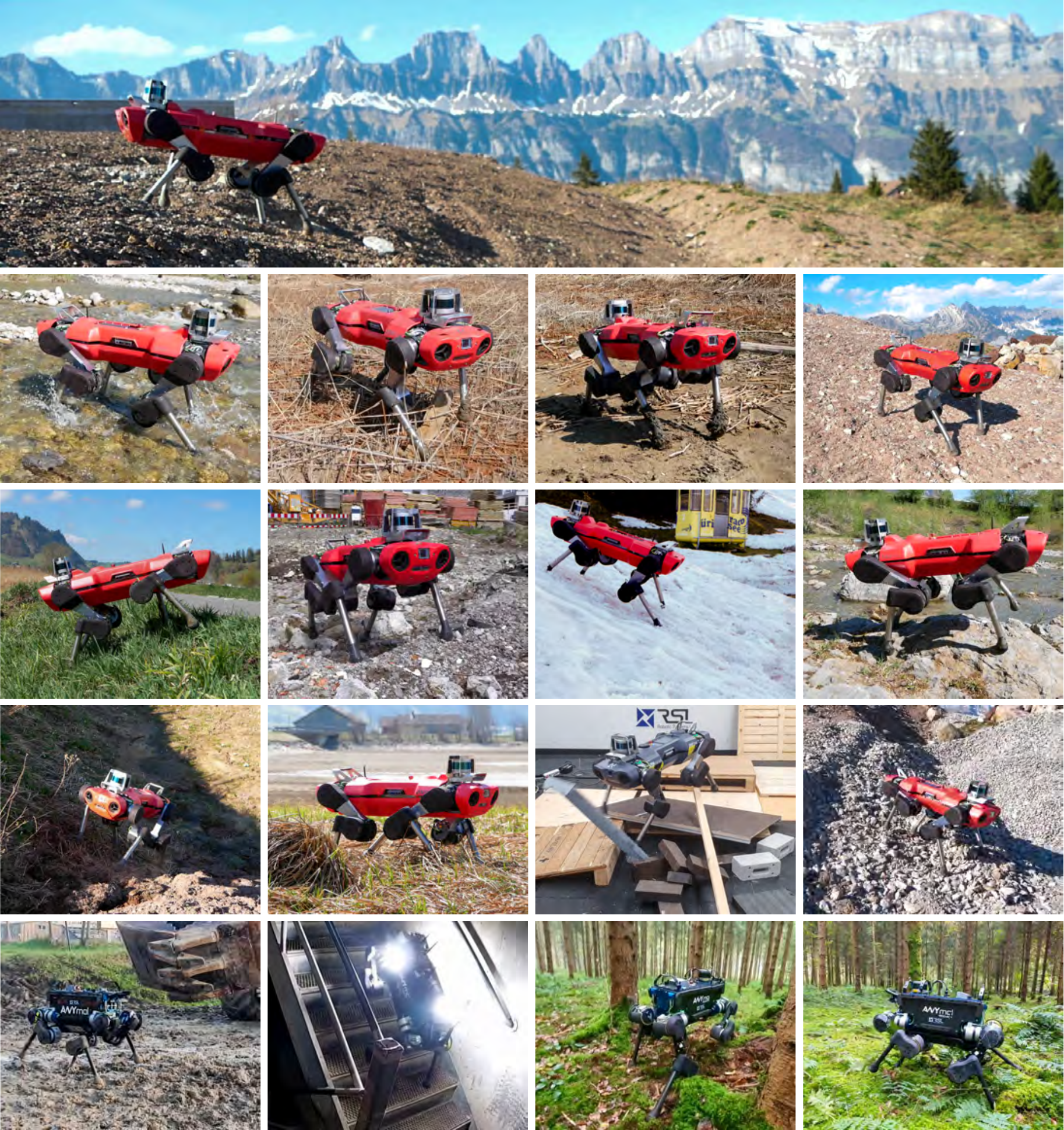}
    \caption{\textbf{Deployment of the presented locomotion controller in a variety of challenging environments.}}
    \label{fig:1}
\end{figure*}

\section{Introduction}
Legged locomotion can dramatically expand the reach of robotics.
Much of the dry landmass on Earth remains impassible to wheeled and tracked machines, the stability of which can be severely compromised on challenging terrain.
Quadrupedal animals, on the other hand, can access some of the most remote parts of our planet. 
They can choose safe footholds within their kinematic reach and rapidly change their kinematic state in response to the environment. Legged robots have the potential to traverse any terrain that their animal counterparts can.

To date, no published work has demonstrated dynamic locomotion in diverse, challenging natural environments as shown in Fig.~\ref{fig:1}.
These environments have highly irregular profiles, deformable terrain, slippery surfaces, and overground obstructions.
Under such conditions, existing published controllers manifest frequent foot slippage, loss of balance, and ultimately catastrophic failure.
The challenge is exacerbated by the inaccessibility of veridical information about the physical properties of the terrain.
Exteroceptive sensors such as cameras and LiDAR cannot reliably measure physical characteristics such as friction and compliance, are impeded by obstructions such as vegetation, snow, and water, and may not have the coverage and temporal resolution to capture changes induced by the robot itself, such as the crumbling of loose ground under the robot's feet.
Under these conditions, the robot must rely crucially on proprioception~-- the sensing of its own bodily configuration at high temporal resolution.
In response to unforeseen events such as unexpected ground contact, terrain deformation, and foot slippage, the controller must rapidly produce whole-body trajectories subject to multiple objectives: balancing, avoiding self-collision, counteracting external disturbances, and locomotion. While animals instinctively solve this complex control problem, it is an open challenge in robotics.

Conventional approaches to legged locomotion on uneven terrain have yielded increasingly complex control architectures. Many rely on elaborate state machines that coordinate the execution of motion primitives and reflex controllers~\cite{jenelten2019dynamic,bledt2018contact,focchi2020heuristic,Reher2019,gong2019feedback}. To trigger transitions between states or the execution of a reflex, many systems explicitly estimate states such as ground contact and slippage~\cite{hwangbo2016probabilistic,camurri2017probabilistic,focchi2018slip}. Such estimation is commonly based on empirically tuned thresholds and can become erratic in the presence of unmodeled factors such as mud, snow, or vegetation. Other systems employ contact sensors at the feet, which can become unreliable in field conditions~\cite{Bloesch2013,gehring2015dynamic,hartley2018legged}. Overall, conventional systems for legged locomotion on rough terrain escalate in complexity as more scenarios are taken into account, have become extremely laborious to develop and maintain, and remain vulnerable to corner cases.

Model-free reinforcement learning (RL) has recently emerged as an alternative approach in the development of legged locomotion skills~\cite{hwangbo2019learning,haarnoja2018learning,xie2019iterative}.
The idea of RL is to tune a controller to optimize a given reward function. The optimization is performed on data acquired by executing the controller itself, which improves with experience. RL has been used to simplify the design of locomotion controllers, automate parts of the design process, and learn behaviors that could not be engineered with prior approaches~\cite{hwangbo2019learning,lee2019robust,haarnoja2018learning,xie2019iterative}.

However, application of RL to legged locomotion has largely been confined to laboratory environments and conditions. Our prior work demonstrated end-to-end learning of locomotion and recovery behaviors~-- but only on flat ground, in the lab~\cite{hwangbo2019learning}. Other work also developed RL techniques for legged locomotion, but likewise focused largely on flat or moderately textured surfaces in laboratory settings~\cite{tan2018sim,haarnoja2018learning,xie2019iterative,yang2019data,ha2020learning,Peng2020}.

Here we present a radically robust controller for blind quadrupedal locomotion on challenging terrain. The controller uses only proprioceptive measurements from joint encoders and an inertial measurement unit (IMU), which are the most durable and reliable sensors on legged machines. The operation of the controller is shown in Fig.~\ref{fig:1} and \href{https://youtu.be/9j2a1oAHDL8}{Movie~1}. The controller was used to drive two generations of ANYmal quadrupeds~\cite{hutter2016anymal} in a variety of conditions that are beyond the reach of prior published work in legged robotics. The controller reliably trots through mud, sand, rubble, thick vegetation, snow, running water, and a variety of other off-road terrain. The same controller was also used in our entry in the DARPA Subterranean Challenge Urban Circuit.
In all deployments, robots of the same generation were driven by exactly the same controller under all conditions.
No tuning was required to adapt to different environments.

Like a number of prior applications of model-free RL to legged locomotion, we train the controller in simulation~\cite{hwangbo2019learning,tan2018sim,xie2019iterative}. Prior efforts have established a number of practices for successful transfer of legged locomotion controllers from simulation to physical machines. One is realistic modeling of the physical system, including the actuators~\cite{hwangbo2019learning}. Another is randomization of physical parameters that vary between simulation and reality, such that the controller becomes robust to a range of conditions that cover those that arise in physical deployment, without the necessity to precisely model these conditions a priori~\cite{peng2018sim}.

We use these ideas as well, but have found that they were not sufficient to achieve robust locomotion on rough terrain. We therefore introduce and validate a number of additional ingredients that are crucial to realizing the presented skills. The first is a different policy architecture. Rather than using a multi-layer perceptron (MLP) that operates on a snapshot of the robot's current state, as was common in prior work, we use a sequence model, specifically a temporal convolutional network (TCN)~\cite{bai2018empirical} that produces actuation based on an extended history of proprioceptive states. We do not employ explicit contact and slip estimation modules, which are known to be brittle in challenging situations; rather, the TCN learns to implicitly reason about contact and slippage events from proprioceptive history as needed.

The second important idea that enables the demonstrated results is privileged learning~\cite{chenlearning}. We have found that training a rough-terrain locomotion policy directly via reinforcement learning was not successful: the supervisory signal was sparse and the presented network failed to learn locomotion within reasonable time budgets. Instead, we decompose the training process into two stages. First, we train a teacher policy that has access to privileged information, namely ground-truth knowledge of the terrain and the robot's contact with it. The privileged information enables the policy to quickly achieve high performance. We then use this privileged teacher to guide the learning of a purely proprioceptive student controller that only uses sensors that are available on the real robot. This privileged learning protocol is enabled by simulation, but the resulting proprioceptive policy is not confined to simulation and is deployed on physical machines.

The third idea that has proven important in achieving the presented levels of robustness is an automated curriculum that synthesizes terrains adaptively, based on the controller's performance at different stages of the training process. In essence, terrains are synthesized such that the controller is capable of traversing them while becoming more robust. We evaluate the traversability of parameterized terrains and use particle filtering to maintain a distribution of terrain parameters of medium difficulty~\cite{brant2017minimal,wang2019paired} that adapt as the neural network learns. The training conditions grow increasingly more challenging, yielding an omnidirectional controller that combines agility with unprecedented resilience.

The result is a legged locomotion controller that is far more robust than any counterparts in existing literature. Remarkably, the controller is consistently effective in \emph{zero-shot generalization} settings. That is, it remains robust when tested in conditions that have never been encountered during training. Our training in simulation only uses rigid terrains and a small set of procedurally generated terrain profiles, such as hills and steps. Yet when deployed on physical quadrupeds, the controller successfully handles deformable terrain (mud, moss, snow), dynamic footholds (stepping on a rolling board in a cluttered indoor environment, or debris in the field), and overground impediments such as thick vegetation, rubble, and gushing water. Our methodology and results open new frontiers for legged robotics and suggest that the extraordinary complexity of the physical world can be tamed without brittle and painstaking modeling or dangerous and expensive trial-and-error in field conditions.

\section{Results}
\begin{figure}
    \centering
  \includegraphics[width=\linewidth]{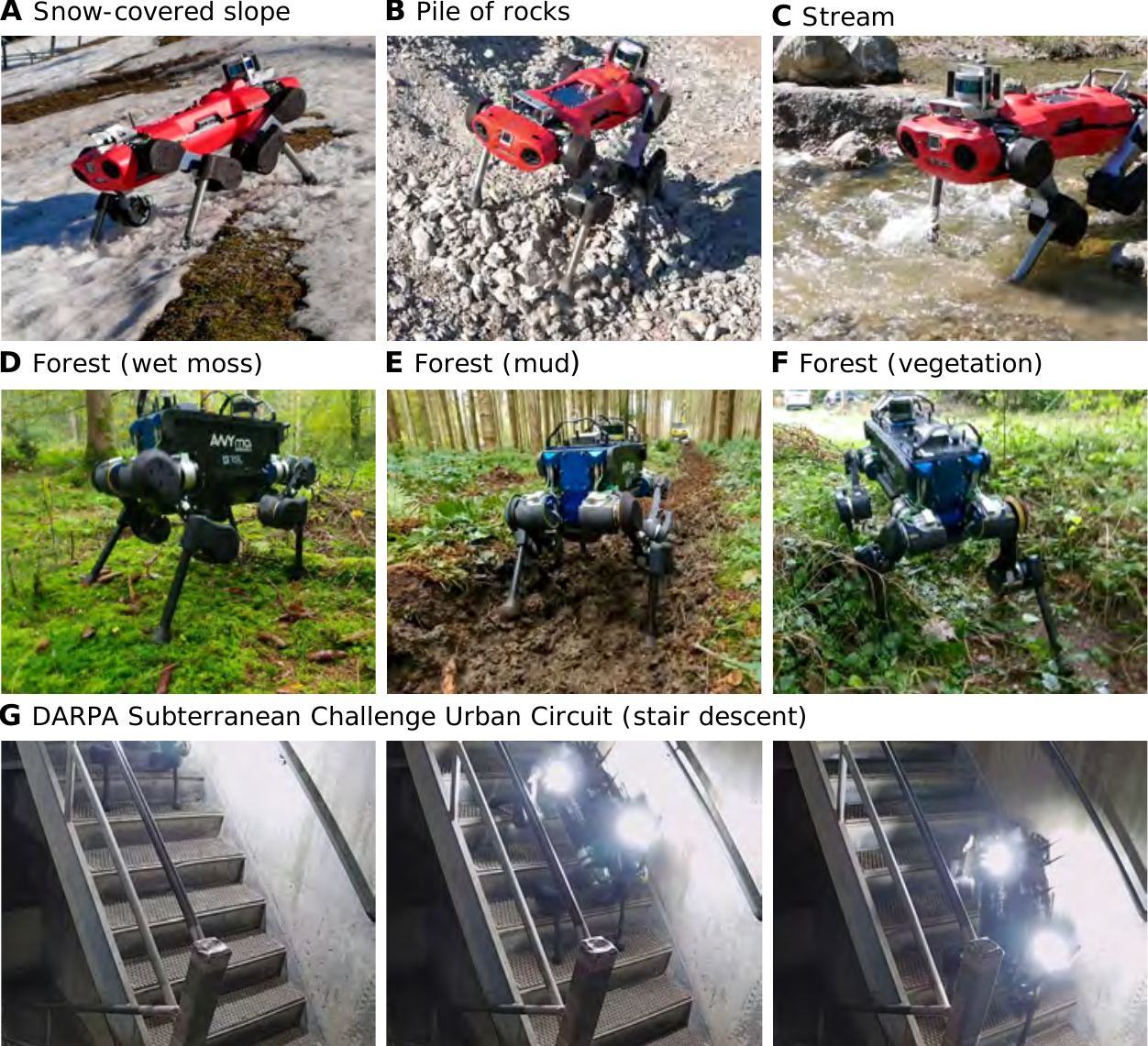}
    \caption{\textbf{A number of specific deployments.}
    (\textbf{A-F}) Zero-shot generalization to slippery and deforming terrain.
    (\textbf{G}) Steep descent during the DARPA Subterranean Challenge. The stair rise is 18 cm and the slope is $\sim 45^{\circ}$.
      }\label{fig:2}
\end{figure}
\begin{figure*}
    \centering
     \makebox[\textwidth][c]{
    \includegraphics[width=\textwidth]{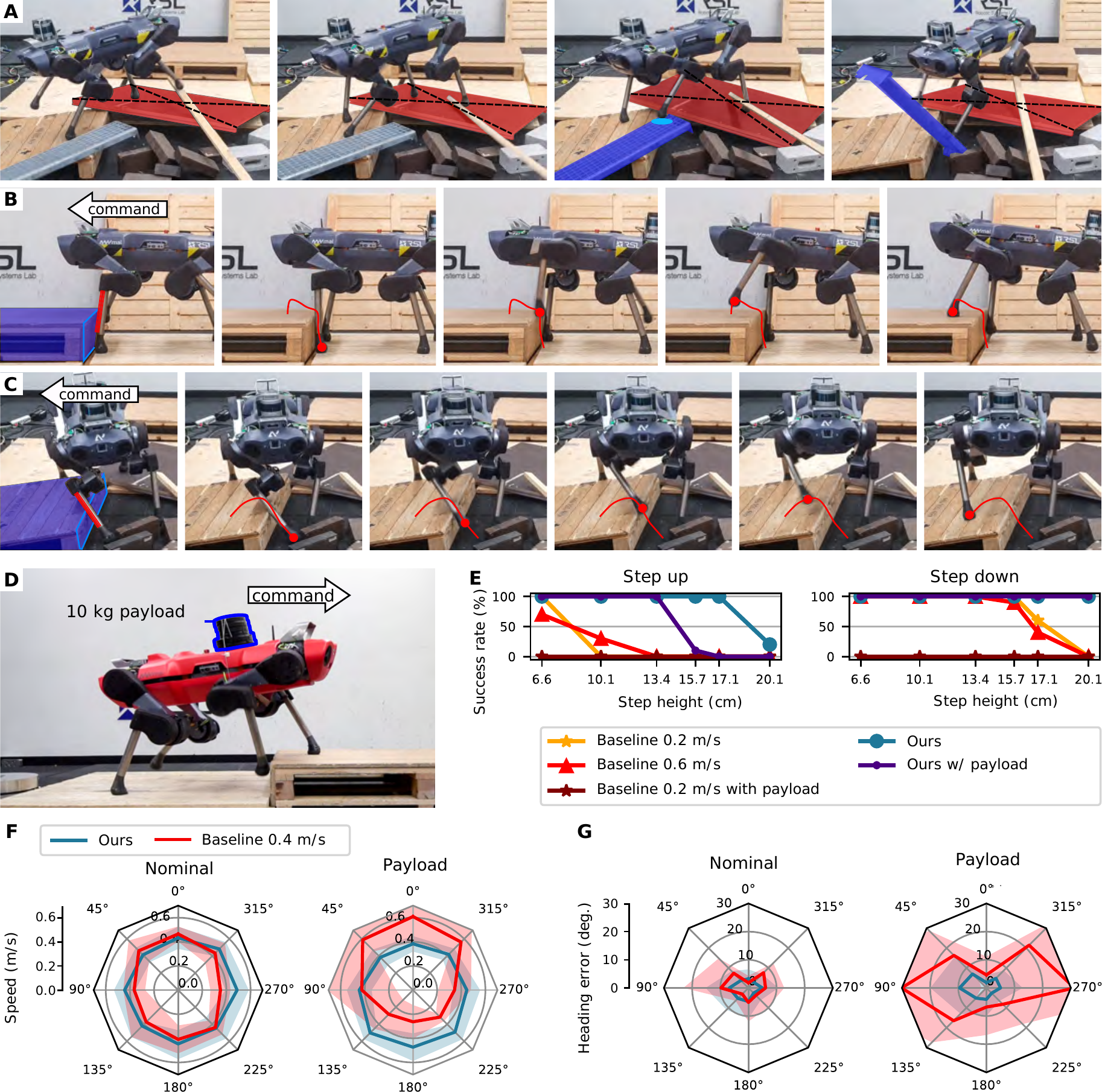}
    }
    \caption{\textbf{Evaluation in an indoor environment.}
    (\textbf{A}) Locomotion over unstable debris. The robot steps onto loose boards (highlighted in red and blue) that dislodge under the robot's feet.
    (\textbf{B}) The policy exhibits a foot-trapping reflex and overcomes a \unit[16.8]{cm} step.
    (\textbf{C}) The policy learns to appropriately handle obstructions irrespective of the contact location. Here it is shown reacting to an obstacle that is encountered mid-shin during the swing phase.
    (\textbf{D}) Controlled experiments with steps and payload. Our controller and a baseline~\cite{bellicoso2018dynamic,jenelten2019dynamic} are commanded to walk over a step with and without the \unit[10]{kg} payload.
    (\textbf{E}) Success rates for different step heights. The success rate is evaluated over 10 trials for each condition.
    (\textbf{F}) Mean linear speed for different command directions on flat terrain. \unit[0]{$^\circ$}  refers to the front of the robot. Shaded area denotes \unit[95]{\%} confidence interval. (\textbf{G}) Mean heading errors for different command directions on flat terrain. Shaded area denotes \unit[95]{\%} confidence interval.
    }\label{fig:3}
\end{figure*}

\href{https://youtu.be/9j2a1oAHDL8}{Movie~1} summarizes the results of the presented work.
We have deployed the trained locomotion controller on two generations of ANYmal robots: ANYmal-B (Fig.~\ref{fig:2}D-G) and ANYmal-C (Fig.~\ref{fig:2}A-C and Fig.~\ref{fig:3}).
The robots have different kinematics, inertia, and actuators.

\subsection*{Natural environments}

The presented controller has been deployed in diverse natural environments, as shown in Fig.~\ref{fig:1} and \href{https://youtu.be/9j2a1oAHDL8}{Movies 1} and \href{https://youtu.be/txjqn8h6pjU}{S1}. These include steep mountain trails, creeks with running water, mud, thick vegetation, loose rubble, snow-covered hills, and a damp forest.
A number of specific scenarios are further highlighted in Fig.~\ref{fig:2}A-F. These environments have characteristics that the policy does not experience during training. The terrains can deform and crumble, with significant variation of material properties over the surface. The robot's legs are subjected to frequent disturbances due to vegetation, rubble, and sticky mud.
Existing terrain estimation pipelines that use cameras or LiDAR~\cite{fankhauser2014robot} fail in environments with snow (Fig.~\ref{fig:2}A), water (Fig.~\ref{fig:2}C), or dense vegetation (Fig.~\ref{fig:2}F). Our controller does not rely on exteroception and is immune to such failure. The controller learns omnidirectional locomotion based on a history of proprioceptive observations and is robust in zero-shot deployment on terrains with characteristics that were never experienced during training.

We have compared the presented controller to a state-of-the-art baseline~\cite{bellicoso2018dynamic,jenelten2019dynamic} in the forest environment. The baseline could traverse flat and unobstructed patches, but failed frequently upon encountering loose branches, thick vegetation, and mud, as shown in \href{https://youtu.be/txjqn8h6pjU}{Movie~S1}. Our controller never failed in these experiments.

\begin{table}
\centering
 \begin{tabular}{|c|c|c|c|c|}
\hline 
\multirow{2}{*}{Quantity} & \multirow{2}{*}{Controller} &  \multicolumn{3}{|c|}{Terrain} \\ \cline{3-5}
  & & Moss & Mud & Vegetation \\ \hline
  \multirow{2}{*}{ \parbox{2cm}{\centering Average speed \\ ($m/s$)}} & 
Ours & \textbf{0.452} & \textbf{0.338} & \textbf{0.248} \\ \cline{2-5}
& Baseline & 0.199 & 0.197 & --  \\ \hline
  \multirow{2}{*}{ \parbox{2cm}{\centering Average \\ mechanical \\ COT}} & 
Ours & \textbf{0.423} & \textbf{0.692} & \textbf{1.23} \\ \cline{2-5}
& Baseline & 0.625 & 0.931& --  \\ \hline
 \end{tabular}
\caption{\textbf{Comparison of locomotion performance in natural environments.} The mechanical COT is computed using positive mechanical power exerted by the actuators.}
\label{tab:1}
\end{table}

We have quantitatively evaluated the presented controller and the baseline in three conditions: moss, mud, and vegetation (Fig.~\ref{fig:2}D-F). We have measured locomotion speed and energy efficiency. The results are reported in Table~\ref{tab:1}. The presented controller achieves higher locomotion speed in all conditions.
We computed the dimensionless cost of transport (COT) to compare the efficiency of the controllers at different speed ranges. 
We define mechanical COT as $\sum_{\text{12 actuators}}{[\tau \dot{\theta}]^{+}}  / (mgv)$. $\tau$ denotes joint torque, $\dot{\theta}$ is joint speed, $mg$ is the total weight, and $v$ is the locomotion speed.
This quantity represents positive mechanical power exerted by the actuator per unit weight and unit locomotion speed~\cite{collins2005efficient}.
As shown in Table~\ref{tab:1}, the presented controller is more energy-efficient, with a lower COT than the baseline.

The quantitative evaluation reported in Table~\ref{tab:1} understates the difference between the two controllers because it only measures speed and energetic efficiency of the baseline when it successfully locomotes.
The baseline's catastrophic failures are not factored into these measurements: when the baseline fails, it is reset by a human operator in a more stable configuration. Catastrophic failures of the baseline controller due to thick vegetation and other factors are shown in \href{https://youtu.be/txjqn8h6pjU}{Movie~S1}. Our controller exhibited no such failures.

\subsection*{DARPA Subterranean Challenge}
Our controller was used by the CERBERUS team for the DARPA Subterranean Challenge Urban Circuit (Fig.~\ref{fig:2}G). It replaced a model-based controller that had been employed by the team in the past~\cite{bellicoso2018dynamic,jenelten2019dynamic}.
The objective of the competition is to develop robotic systems that rapidly map, navigate, and search complex underground environments, including tunnels, urban underground, and cave networks.
The human operators are not allowed to assist the robots during the competition physically; only teleoperation is allowed.
Accordingly, the locomotion controller needs to perform without failure over extended mission durations.
To our knowledge, this is the first use of a legged locomotion controller trained via model-free RL in such competitive field deployment.

The presented controller drove two ANYmal-B robots in four missions of 60 minutes.
The controller exhibited a zero failure rate throughout the competition.
A steep staircase that was traversed by one of the robots during the competition is shown in Fig.~\ref{fig:2}G.

\subsection*{Indoor experiments}
We further evaluated the robustness of the presented controller in an indoor environment populated by loose debris, as shown in Fig.~\ref{fig:3}A.
Support surfaces are unstable and the robot's feet frequently slip.
Such conditions can be found at disaster sites and construction zones, where legged robots are expected to operate in the future.

Results are shown in Fig.~\ref{fig:3}A and \href{https://youtu.be/Xnn4sVSpSh0}{Movie~S2}. The robot moves omnidirectionally over the area. The presented controller can stably locomote over shifting support surfaces.
This level of robustness is beyond the reach of prior controllers for ANYmal robots~\cite{bellicoso2018dynamic,jenelten2019dynamic} and is comparable to the state of the art~\cite{bledt2018contact,vision60blind}.

The learned controller manifests a foot-trapping reflex, as shown in Fig.~\ref{fig:3}B and \href{https://youtu.be/tPixnjLbTvE}{Movie~S3}.
The policy identifies the trapping of the foot purely from proprioceptive observations and lifts the foot over the obstacle.
%than the usual foot clearance (i.e., maximum height of a swing foot).
Such reflexes were not specified in any way during training: they developed adaptively. This distinguishes the presented approach from conventional controller design methods, which explicitly build in such reflexes and orchestrate their execution by a higher-level state machine~\cite{jenelten2019dynamic,focchi2020heuristic}.
The step shown in Fig.~\ref{fig:3}B is \unit[16.8]{cm} high, which is higher than the foot clearance of the legs during normal walking on flat terrain.
The maximum foot clearance on flat terrain is \unit[12.9]{cm} and \unit[13.6]{cm} for the LF and RF legs, respectively\footnote{ We denote left, right, fore, and hind as L, R, F, H, respectively, to compactly refer to a leg. For example, `LF leg' refers to the left fore leg.}, and increases up to \unit[22.5]{cm} and \unit[18.5]{cm} in the case of foot-trapping.
Our controller also learns to adapt the hind leg trajectories when stepping up. The maximum foot clearance on flat terrains is \unit[13.5]{cm} and \unit[9.06]{cm} for the LH and RH legs, and increases up to \unit[16.6]{cm} and \unit[15.9]{cm} when the front legs are above the step. Further analysis is provided in the Materials and Methods section.
Note also that the reflexes learned by our controller are more general and are not tied to particular contact events. Fig.~\ref{fig:3}C shows the controller responding to a mid-shin collision during the swing phase. Here, the trapping event was not signalled by foot contact, and scripted controllers that use foot contact events as triggers would not appropriately handle this situation. Our controller, on the other hand, analyzes the proprioceptive stream as a whole and is trained without making assumptions about possible contact locations. Hence, it can learn to react to any obstructions and disturbances that impact the robot's bodily configuration.

%\subsection*{Controlled comparison with the existing method}
We now focus on comparing the presented approach with the baseline~\cite{bellicoso2018dynamic,jenelten2019dynamic} in controlled settings.
We first compare the robustness of the controllers in the diagnostic setting of a single step, as shown in Fig.~\ref{fig:3}D.
In each trial, the robot is driven straight to a step for \unit[10]{s}. A trial is a success if the robot traverses the step with both front and hind legs. We conducted 10 trials for each step height and computed the success rate.
Since the baseline controller takes a desired linear velocity of the base as input, we commanded a forward velocity of \unit[0.2]{m/s} and \unit[0.6]{m/s}. \unit[0.6]{m/s} is the maximum speed of the baseline.
The success rates are given in Fig.~\ref{fig:3}E. The presented controller outperforms the baseline in both stepping up and down.
The baseline showed high sensitivity to foot-trapping, which often led to a fall, as shown in \href{https://youtu.be/tPixnjLbTvE}{Movie~S3}.

We also tested the controllers in the presence of substantial model mismatch. We attached a \unit[10]{kg} payload, as shown in Fig.~\ref{fig:3}D and \href{https://youtu.be/3Nr47MXCFO0}{Movie~S4}. This payload is \unit[22.7]{\%} of the total weight of the robot, and was never simulated during training. As shown in Fig.~\ref{fig:3}E, the presented controller can still traverse steps up to \unit[13.4]{cm} despite the model mismatch. The baseline is incapable of traversing any steps under any command speed with the payload.

We then evaluate the tracking performance of the controllers on flat ground with the payload.
We commanded each controller in 8 directions and measured the locomotion speed and the tracking error.
Target speed is fixed to \unit[0.4]{m/s} for the baseline controller, which is similar to the operating speed of the presented controller.
In Fig.~\ref{fig:3}F, we show the velocity profiles of the controllers. Our controller locomotes at around \unit[0.4]{m/s} in all directions and performs similarly with the payload.
On the other hand, the locomotion speed of the baseline varies with direction, which can be seen by the anisotropic velocity profile, and the velocity profile shifts significantly off center with the payload.
Fig.~\ref{fig:3}G shows the heading error of the controllers in each commanded direction. The heading error is the angle between the command velocity and the base velocity of the robot. The heading error of the presented controller is consistently smaller than the baseline, both with and without the payload.
The baseline's error in the lateral direction reaches $\sim$\unit[30]{$^\circ$} and the baseline fails when a speed of (\unit[0.6]{m/s}) is commanded, as shown in \href{https://youtu.be/3Nr47MXCFO0}{Movie~S4}. 
In contrast, the average heading error of the presented controller stays within \unit[10]{$^\circ$} with or without the payload. We conclude that the presented controller is much more robust to model mismatch.

Next we test robustness to foot slippage.
To introduce slippage, we used a moistened whiteboard~\cite{jenelten2019dynamic}.
The results are shown in \href{https://youtu.be/aMPwB3t4idU}{Movie~S5}.
The baseline quickly loses balance, aggressively swings the legs, and falls. In contrast, the presented controller adapts to the slippery terrain and successfully locomotes in the commanded direction.

\section{Discussion}
The presented results substantially advance the published state of the art in legged robotics. Beyond the results themselves, the methodology presented in this work can have broad applications. Prior to our work, a hypothesis could be held that training in simulation is fundamentally constrained by the limitations of simulation environments in representing the complexity of the physical world. Present-day technology is severely limited in its ability to simulate compliant contact, slippage, and deformable and crumbling terrain. As a result, phenomena such as mud, snow, thick vegetation, gushing water, and many others are beyond the capabilities of robotics simulation frameworks~\cite{hwangbo2018per,coumans2013bullet,smith2005open}. The sample complexity of model-free RL algorithms, which commonly require millions of time steps for training, further exacerbates the challenge by precluding reliance on frameworks that may require seconds of computation per time step.

Our work demonstrates that simulating the astonishing variety of the physical world may not be necessary. Our training environment features only rigid terrain, with no compliance or overground obstructions such as vegetation. Nevertheless, controllers trained in this environment successfully meet the diversity of field conditions encountered at deployment.

We see a number of limitations and opportunities for future work. First, the presented controller only exhibits the trot gait. This is narrower than the range of gate patterns discovered by quadrupeds in nature~\cite{Alexander2003}. The gait pattern is constrained in part by the kinematics and dynamics of the robot, but the ANYmal machines are physically capable of multiple gates~\cite{bellicoso2018dynamic}. We hypothesize that training protocols and objectives that emphasize diversity can elicit these.

Second, the presented controller relies solely on proprioception. This is a significant advantage in that the controller makes few assumptions on the sensor suite and is not susceptible to failure when exteroception breaks down. Indeed, existing work has argued that a blind (proprioceptive) controller should form the basis of a legged locomotion stack~\cite{focchi2020heuristic}. Nevertheless, blind locomotion is inherently limited. If the machine is commanded to walk off a cliff, it will. Even in less extreme conditions, the robot's gait is fairly conservative since it must by necessity feel out the environment with its body as it locomotes. A major opportunity for future work is to use the presented methodology as a starting point in the development of a hybrid proprioceptive-exteroceptive controller that, like many animals, will be able to locomote even when vision and other external senses are disrupted, but will use exteroceptive data when it is provided. This will enable legged machines to autonomously traverse environments that may have fatal elements such as cliffs, and to raise speed and energetic efficiency in safer conditions.

More broadly, the presented results expedite the deployment of legged machines in environments that are beyond the reach of wheeled and tracked robots and are dangerous or inaccessible to humans, while the presented methodology opens new frontiers for training complex robotic systems in simulation and deploying them in the full richness and complexity of the physical world.

\section{Materials and Methods}

\begin{figure*}
    \centering
    \includegraphics[width=\textwidth]{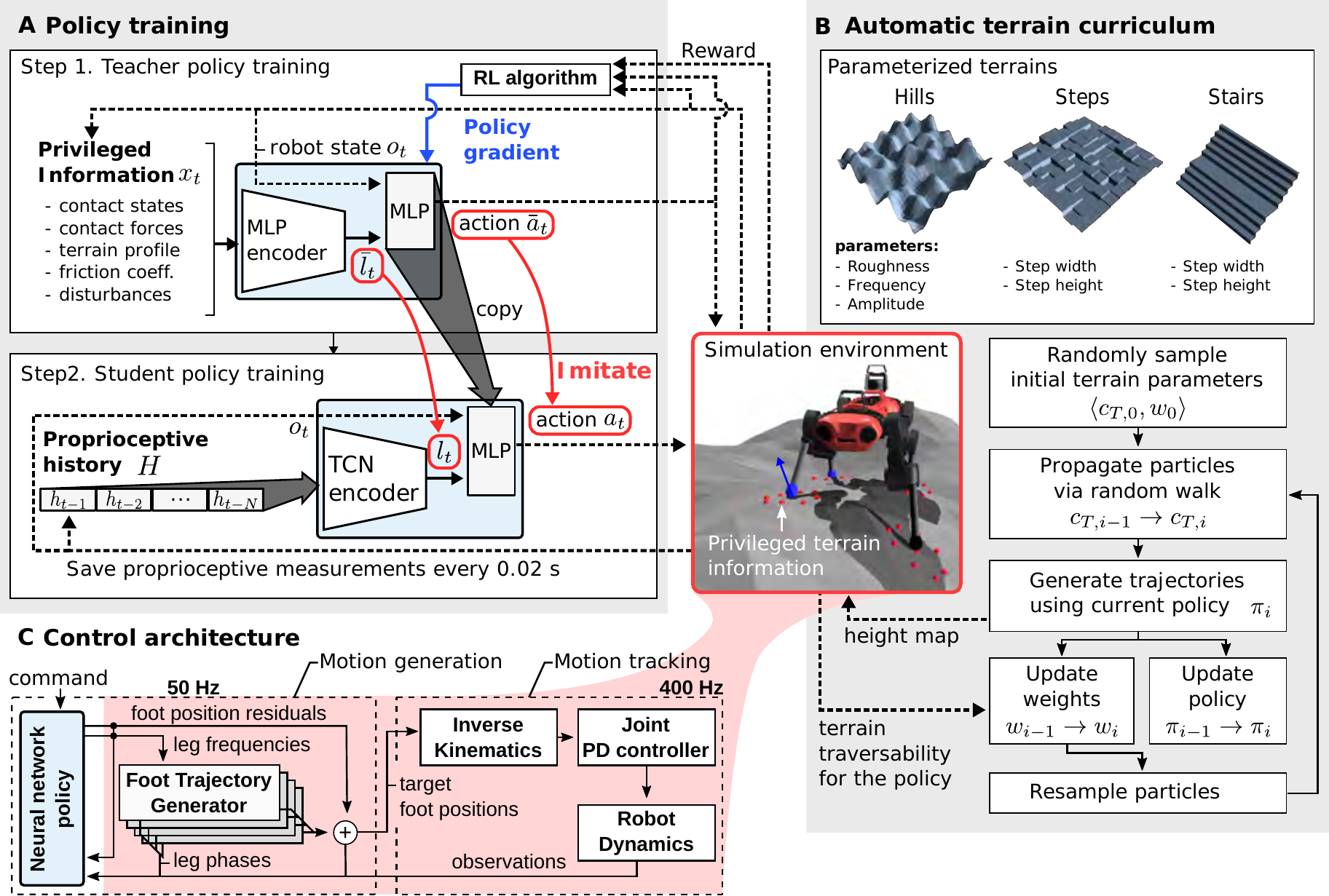}
    \caption{\textbf{Overview of the presented approach.} (\textbf{A})~Two-stage training process. First, a teacher policy is trained using reinforcement learning in simulation. It has access to privileged information that is not available in the real world.
    Next, a proprioceptive student policy learns by imitating the teacher. The student policy acts on a stream of proprioceptive sensory input and does not use privileged information.
    (\textbf{B})~An adaptive terrain curriculum synthesizes terrains at an appropriate level of difficulty during the course of training. Particle filtering is used to maintain a distribution of terrain parameters that are challenging but traversable by the policy.
    (\textbf{C})~Architecture of the locomotion controller. The learned proprioceptive policy modulates motion primitives via kinematic residuals. An empirical model of the joint PD controller facilitates deployment on physical machines.
    }
    \label{fig:4}
\end{figure*}

\subsection{Overview}

The main objective of the presented controller is to locomote over rough terrain following a command.
The command is given either by a human operator or by a higher-level navigation controller.
In our formulation, unlike many existing works~\cite{hwangbo2019learning,tan2018sim,xie2019iterative} that focus on tracking the target velocity of the base ($^B_{IB} v_T$), only the direction ($^B_{IB} \hat{v}_T$) is given to the controller.
The reason is that the feasible range of target speeds is often unclear on challenging terrain. For example, the robot can walk faster downhill than uphill.

The command vector is defined as $\langle (^{B}_{IB}\hat{v}_{T})_{xy}, (\hat{\omega}_{T})_{z} \rangle$.
The first part is the target horizontal direction in base frame $(^{B}_{IB}\hat{v}_{T})_{xy} \coloneqq \langle \cos(\psi_T), \sin(\psi_T) \rangle$, where $\psi_T$ is the yaw angle to command direction in the base frame. The stop command is defined as $\langle0.0, 0.0\rangle$. The second part is the turning direction $(\hat{\omega}_{T})_{z} \in \{-1, 0, 1 \}$. 1 refers to counter-clockwise rotation along the base $z$-axis.

An overview of our method is given in Fig.~\ref{fig:4}. 
We use a privileged learning strategy inspired by ``learning by cheating''~\cite{chenlearning} (Fig.~\ref{fig:4}A).
We first train a teacher policy that has access to privileged information concerning the terrain. This teacher policy is then distilled into a proprioceptive student policy that does not rely on privileged information. The privileged teacher policy is confined to simulation, but the student policy is deployed on physical machines. One difference of our methodology from that of Chen et al.~\cite{chenlearning} is that we do not rely on expert demonstrations to train the privileged policy; rather, the teacher policy is trained via reinforcement learning.

The privileged teacher model is based on multi-layer perceptrons (MLPs) that receive information on the current state of the robot, properties of the terrain, and the robot's contact with the terrain. The model computes a latent embedding $\bar{l}_t$ that represents the current state, and an action $\bar{a}_t$.
The training objective rewards locomotion in prescribed directions.

After the teacher policy is trained, it is used to supervise a proprioceptive student policy. The student model is a temporal convolutional network (TCN)~\cite{bai2018empirical} that receives a sequence of $N$ proprioceptive observations as input. The student policy is trained by imitation. The vectors $\bar{l}_t$ and $\bar{a}_t$ computed by the teacher policy are used to supervise the student. This is illustrated in Fig.~\ref{fig:4}A.

Training is conducted on procedurally generated terrains in simulation. The terrains are synthesized adaptively, to facilitate learning according to the skill level of the trained policies at any given time. We define a traversability measure for terrain and develop a sampling-based method to select terrains with the appropriate difficulty during the course of training. We use particle filtering to maintain an appropriate distribution of terrain parameters. This is illustrated in Fig.~\ref{fig:4}B.
The terrain curriculum is applied during both teacher and student training.

Our control architecture is shown in Fig.~\ref{fig:4}C.
We employ the Policies Modulating Trajectory Generators (PMTG) architecture~\cite{iscen2018policies} to provide priors on motion generation.
The neural network policy modulates leg phases and motion primitives by synthesizing residual position commands.

The simulation uses a learned dynamics model of the robot's joint PD controller~\cite{hwangbo2019learning}. This facilitates the transfer of policies from simulation to reality. After training in simulation, the proprioceptive controller is deployed directly on physical legged machines, with no fine-tuning.

\subsection*{Motion synthesis}

We now elaborate on the control architecture that is illustrated in Fig.~\ref{fig:4}C. It is divided into motion generation and tracking.
The input to our controller consists of the command vector and a sequence of proprioceptive measurements including base velocity, orientation and joint states.
The controller does not use any exteroceptive input (e.g., no haptic sensors, cameras, or depth sensors). The input also does not contain any handcrafted features such as foot contact states or estimated terrain geometry.
The controller outputs joint position targets. 

Our motion generation strategy is based on the periodic leg phase.
Previous works commonly leveraged predefined foot contact schedules~\cite{barasuol2013reactive,bledt2018contact,bellicoso2018dynamic}.
We define a periodic phase variable $\phi_{i} \in [0.0, 2\pi)$ for each leg, which represents contact phase if $\phi \in [0.0, \pi)$ and swing phase if $\phi \in [\pi, 2\pi)$.
At every time step $t$, $\phi_i = (\phi_{i,0} + (f_0 + f_i)t ) \pmod{2 \pi}$ where $\phi_{i,0}$ is the initial phase, $f_0$ is a common base frequency, and $f_i$ is the frequency offset for the $i$-th leg. We want the legs to manifest periodic motions when $f_0+f_i \neq 0$ and engage ground contact in contact phase.
We set $f_0$ as \unit[1.25]{Hz}, which is the value used by a previously developed conventional controller for a trot gait~\cite{bellicoso2018dynamic}.

The target foot positions, which are the output of the motion generation block, are defined in the horizontal frames~\cite{barasuol2013reactive} of the feet ($H_{i}, i\in \{1,2,3,4 \}$).
$H_i$ is a reference frame that is attached below the hip joint of the $i$-th leg.
The distance equals the nominal reach of the leg.
The $z$-axis of the frame ($^{H_i}z$) is parallel to $e_g$ and $^{H_i}x$ is the projection of the base $x$-axis ($^{B}x$) onto the horizontal plane, i.e., the frame has the same yaw angle with the robot.
The roll and pitch angles of $H_i$ are decoupled from the base.
This kinematic trick reduces the effect of base attitude on the foot motions~\cite{barasuol2013reactive} and consequently stabilizes training.
Defining the output in $H_i$ results in less premature termination at the beginning of the policy training, when the base motion is unstable due to random actions.
Another benefit is that we can decompose the action distribution of the stochastic policy in the lateral and vertical directions during policy training. We applied larger noise in the lateral direction to promote exploration along the ground surface.

We use the PMTG architecture~\cite{iscen2018policies} to integrate a neural network to regulate the controller. 
Our implementation consists of four identical foot trajectory generators (FTGs) and a neural network policy.
The FTG is a function $F(\phi): [0.0, 2\pi) \rightarrow \mathbb{R}^3$ that outputs foot position targets for each leg. The FTG drives vertical stepping motion when $f_i$ is non-zero. The definition of $F(\phi)$ is given in supplementary section~S3.

The policy outputs $f_i$s and target foot position residuals ($\Delta {r}_{f_i,T}$), and the target foot position for the $i$-th foot is $r_{f_i,T} \coloneqq F(\phi_i) + \Delta{r}_{f_i,T}$.

The tracking control is done using analytic inverse kinematics (IK) and joint position control.
Each foot position target defined in $H_i$ is first expressed in the robot base frame, and the joint position targets are computed using analytic IK. 
The joint position targets are then tracked by joint position PD controllers.
The main reason for using analytic IK is to maximize computational efficiency and to reuse existing position control actuator models~\cite{hwangbo2016probabilistic,lee2019robust} for the sim-to-real transfer.

\subsection*{Teacher policy}

We formulate the control problem as a Markov Decision Process (MDP).
MDP is a mathematical framework for modeling discrete-time control processes in which the evolution of the state and the outcomes are partly stochastic.
An MDP is defined by a state space $\mathcal{S}$, action space $\mathcal{A}$, a scalar reward function $\mathcal{R}(s_t, s_{t+1})$, and the transition probability $P(s_{t+1} | s_t, a_t)$.
A learning agent selects an action $a_t$ from its policy $\pi(a_t|s_t)$ and receives a reward $r_t$ from the environment.
The objective of the RL framework is to find an optimal policy $\pi^*$ that maximizes the discounted sum of rewards over an infinite time horizon.

Assuming the environment is fully observable to the teacher, we formulate locomotion control as an MDP and use an off-the-shelf RL method~\cite{schulman2015trust} to solve it. In this section, we provide the MDP for teacher training, which is defined by a tuple of state space, action space, transition probability, and reward function.

The state is defined as $s_t \coloneqq \langle o_t, x_t \rangle$, where $o_t$ is the measurement vector obtainable from the robot and $x_t$ is the privileged information that is usually not available in the real world. The detailed definitions are given in Table~S4.
$o_t$ contains command, orientation, base twist, joint positions and velocities, $\phi_i$s, $f_i$s, and previous foot position targets. Joint position errors and velocities measured at \unit[-0.01]{s} and \unit[-0.02]{s} are contained in $o_t$, which is the same as the input to the learned model of the joint-level PD controller.
This information allows the policy to exploit the actuator dynamics~\cite{hwangbo2019learning}.
To encode the leg phase, we use $\langle \cos(\phi), \sin(\phi) \rangle$ instead of $\phi$, which is a smooth and unique representation for the angle.
Previous foot position targets are also fed back to the policy and are used to compute the target smoothness reward that is explained in the following paragraph.
When the student controller is deployed, the quantities in $o_t$ are replaced with readings from the proprioceptive sensors and the base velocity and orientation are provided by a state estimator~\cite{bloesch2013state}.
$x_t$ contains noiseless information that we receive directly from a physics engine.
$x_t$ mainly consists of information related to foot-ground interactions such as terrain profile, foot contact states and forces, friction coefficients, and external disturbance forces applied during training. Specifically, we represent the terrain profile with the elevation of 9 scan points around each foot, which are symmetrically placed along a circle with a \unit[10]{cm} radius (visualized in Fig.~\ref{fig:4}).

The action ($\bar{a}_t$) is a 16-dimensional vector consisting of leg frequencies and foot position residuals.

The reward function is defined such that an RL agent receives a higher reward if it advances faster towards the goal. The reward function is specified in detail in supplementary section S4.

The policy network is constructed by two MLP blocks as shown in Fig.~\ref{fig:4}A.
The MLP encoder embeds $x_t$ into a latent vector $\bar{l}_t$.
The command and robot states are not included in $x_t$, so $\bar{l}_t$ contains only the terrain- and contact-related features.
We hypothesize that $\bar{l}_t$ drives adaptive behaviors such as changing foot clearance depending on the terrain profile.
Then $\bar{l}_t$ and $o_t$ are provided to the subsequent MLP layers to compute action.

The Trust Region Policy Optimization (TRPO)~\cite{schulman2015trust} algorithm is used for training.
The hyperparameters we used are given in Table~S7.

\subsection*{Student policy}

The proprioceptive student policy only has access to $o_t$.
A key hypothesis here is that the latent features $\bar{l}_t$ can be (partially) recovered from a time series of proprioceptive observations, $h_t$, which is defined as $ h_t \coloneqq o_t \setminus \{f_o, \text{joint history}, \text{previous foot position targets} \}$.

The student policy uses a temporal convolutional network (TCN)~\cite{bai2018empirical} encoder.
The input to the TCN encoder is $H = \{h_{t-1}, ..., h_{t-N-1}\}$, where $N$ is the history length.
The encoder is fully convolutional and consists of three dilated causal convolutional layers, interleaved with strided convolutional layers that reduce dimensionality.
The architecture is specified in Tables~S5 and~S6.

We use the TCN architecture because it affords transparent control over the input history length, can accommodate long histories, and is known to be robust to hyperparameter settings~\cite{bai2018empirical}. A comparison with a recurrent neural network architecture is provided in supplementary section S8.

The student policy is trained via supervised learning. The loss function is defined as
\begin{equation}
    \mathcal{L} \coloneqq (\bar{a_t}(o_t, x_t) - a_t(o_t, H))^2 +  (\bar{l_t}(o_t, x_t) - l_t(H))^2. 
\label{eq:student}\end{equation}
Quantities marked by a bar ($\bar{\cdot}$) denote target values generated by the teacher. 
We employ the dataset aggregation strategy (DAgger)~\cite{ross2011reduction}. Specifically, training data is generated by rolling out trajectories by the student policy. For each visited state, the teacher policy computes its embedding and action vectors ($\bar{\cdot}$). These outputs of the teacher policy are used as supervisory signals associated with the corresponding states.
The hyperparameters we used are given in Table~S8.

\subsection*{Adaptive terrain curriculum}

Our method is inspired by automatic curriculum learning (ACL) for RL agents~\cite{wang2019paired,florensa2018automatic}. The paired open-ended trailblazer (POET) approach~\cite{wang2019paired} generates diverse parameterized terrains for a 2D bipedal agent.
The method employs minimal criteria (MC)~\cite{lehman2010revising,brant2017minimal} and aims to choose environmental parameters that are neither too challenging nor trivial for the agents: this is realized by selecting task parameters that yield mid-range rewards. Florensa et al.~\cite{florensa2018automatic} similarly choose achievable yet difficult goals for RL agents.

Our method likewise realizes a training curriculum that gradually modifies a distribution over environmental parameters such that the policy can continuously improve locomotion skills and generalize to new environments.
Our work differs from POET as POET aims for open-ended search in the space of possible problems and evolves a population of specialized agents while we seek to obtain a single generalist agent.

Fig.~\ref{fig:4}B shows the types of terrains used in our training environment. Each terrain is generated by a parameter vector $c_T \in \mathcal{C}$. The terrains are described in detail in supplementary section~S5. Our ACL method approximates a distribution of desirable $c_T$s using a particle filter. 

We first describe how a given $c_T$ is evaluated in simulation. Instead of directly using the reward function to evaluate the learning progress~\cite{matiisen2019teacher,wang2019paired,yu2018learning,akkaya2019solving}, we evaluate $c_T$s by the traversability of generated terrains, which is defined as the success rate of traversing a terrain. We found traversability to be more intuitive than the reward function, which consists of multiple objectives that are often unbounded.
We first define a labeling function $\nu$ as
\begin{equation}
  \nu(s_t, a_t, s_{t+1}) = 
\begin{cases}
1  & \text{if}\quad v_{pr}(s_{t+1}) > 0.2\\
0  & \text{if}\quad v_{pr}(s_{t+1}) < 0.2 \lor \text{termination}
\end{cases}{}
~\label{traj_score}
\end{equation}
for a state transition from $s_t$ to $s_{t+1}$. 
$v_{pr}(s_{t+1})$ stands for the inner product of the base velocity and commanded direction at time step $t+1$.
If $\pi$ can locomote in the commanded direction faster than \unit[0.2]{m/s}, we consider the terrain traversable in this direction. The threshold is a hyperparameter; \unit[0.2]{m/s} is about one third of the maximum speed of our robot. Traversability is defined as 
\begin{equation}
Tr(c_T,\pi) = \mathbb{E}_{\xi \sim  \pi} \{ \nu(s_t, a_t, s_{t+1} \mid c_{T}) \} \in [0.0, 1.0],    
\label{def:traversability}\end{equation}
where $\xi$ refers to trajectories generated by $\pi$. 
This follows a definition of empirical traversability in prior work~\cite{chavez2018learning}.

The objective of our terrain generation method is to find $c_T$s with mid-range traversability ($Tr(c_T, \pi)\in [0.5, 0.9]$).
The rationale is to synthesize terrains that are neither too easy nor too difficult. We define terrain desirability as follows:
\begin{align}
    Td(c_T, \pi)  &\coloneqq \Pr  (Tr(c_T,\pi) \in [0.5, 0.9]) \\
    &= \mathbb{E}_{\xi \sim \pi} \{ Tr(c_T,\pi) \in [0.5, 0.9] \},
\label{def:desirability}\end{align}
where 0.5 and 0.9 are fixed thresholds for minimum/maximum traversability.

We use a particle filter to keep track of a distribution of high-desirability $c_T$s during training.
We formulate a particle filtering problem where we approximate the distribution of terrain parameters that satisfies $Tr(c_T,\pi) \in [0.5, 0.9]$ with a finite set of sampling points ($c_{T}^k \in \mathcal{C}, k \in {1, \cdots, N_{particle}}$). 
Our algorithm is modeled on the Sequential Importance Resampling (SIR) particle filter. It is based on the following assumptions.

\begin{enumerate}
    \item Terrain parameters with similar $Tr(\cdot, \pi)$ are close in Euclidean distance in parameter space. 
    \item A policy trained over the terrains generated by $c_T$s in some area of $\mathcal{C}$ will learn to interpolate to nearby terrain parameters.
    \item $c_{T, 0}, c_{T, 1}, ... $ forms a Markov process, where $c_{T,j} = \{c_{T,j}^1, c_{T,j}^2, ...  c_{T,j}^{N_{particle}}\}$ at iteration $j$. 
\end{enumerate}
The first assumption comes from the insight that terrain parameters can be interpolated, e.g., the difficulty of a staircase increases as we increase the step height. The second assumption justifies the use of discrete samples from $\mathcal{C}$ to train a policy that generalizes over a certain region of $\mathcal{C}$. The last assumption is necessary for formulating a particle filter.

The importance weight $w^k$ is defined for each $c_{T}^k$, and the set of tuples $\langle c_{T}^k , w^k \rangle$ approximates the target distribution ($c_T$s with $Tr(c_T,\pi) \in [0.5, 0.9]$).
We define the measurement variable $y_j^k$ such that $y_j^k = 1 $ if $Tr(c^k_{T,j},\pi) \in [0.5, 0.9]$.
Then the terrain desirability defined above becomes the measurement probability 
\begin{equation}
    \Pr(y_j^k|c_{T,j}^k) 
    = \Pr  (Tr(c_{T,i}^k,\pi) \in [0.5, 0.9])
    = Td(c_{T,j}^k,\pi).
\label{measurement_model}\end{equation}
For practical implementation, the measurement probability is computed by the empirical expectation from the samples collected during policy training:
\begin{equation}
    \Pr(y_j^k|c_{T,j}^k) 
    \approx
     \sum_{}^{N_{traj}} 
     \frac{\mathbbm{1} (Tr(c_{T,j}^k,\pi) \in [0.5, 0.9])}{ N_{traj}},
\label{measurement_model2}\end{equation}
where $N_{traj}$ denotes the number of trajectories generated using $c^{k}_{T,j}$. The trajectories are also used for policy training. Our method therefore does not require additional evaluation steps to advance the curriculum of the terrain parameters. Resampling is done such that the probability of choosing the $k$th sample equals the normalized importance weight $w^k/\sum_i^{N_{particle}} w^i \in [0, 1]$. 

The transition model is a random walk in $\mathcal{C}$. Each parameter of a sampling point is shifted to its adjacent value by a fixed probability $p_{transition}$. 
It satisfies the third assumption (Markov process) because the evolution of each parameter only relies on the current value and randomly sampled noise.
To improve exploration, we bounded and discretized $\mathcal{C}$ to reduce the search space. The initial samples ($c_{T,0}^k$) are either drawn uniformly from $\mathcal{C}$ or concentrated on almost flat terrains.

Implementation details and an overview of the training process are provided in supplementary section S2 and Algorithm S1 in the supplement.

\begin{figure*}
    \centering
    \includegraphics[width=\textwidth]{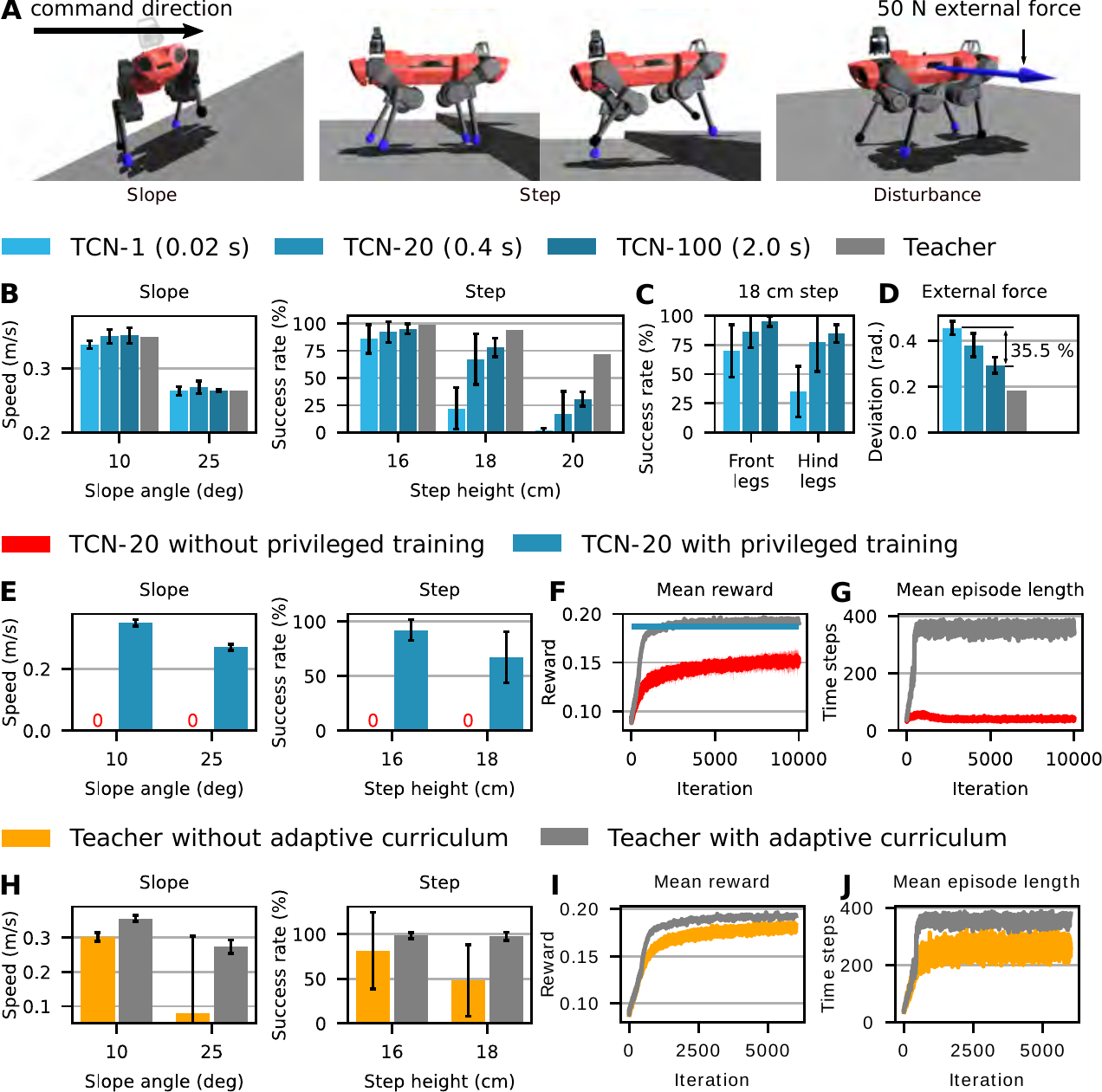}
    \caption{\textbf{Ablation studies.}
    We trained each model 5 times using different random seeds. Error bars denote \unit[95]{\%} confidence intervals.
    (\textbf{A}) Test setups. 
    The robot is commanded to advance for \unit[10]{s} in the specified direction (black arrow).
    We conducted 100 trials for each test.
    On the step test, a trial is considered successful if the robot traverses the step with both front and hind legs. 
    Robots are initialized with random joint configurations.
     Initial yaw angle is sampled from $U(-\pi,\pi)$ for the slope test and from $U(-\pi/6,\pi/6)$ for the other tests. 
    The friction coefficients between the feet and the ground are sampled from $U(0.4,1.0)$. The external force is applied for \unit[5]{s} in the lateral direction.
    (\textbf{B-D}) Importance of memory length $N$ in the TCN-$N$ encoder.
    (\textbf{E-G})~Importance of privileged training.
    (\textbf{F})~Learning curves for the teacher (grey) and a TCN-20 student trained directly, without privileged training (red). For comparison, the blue line indicates the mean reward of a TCN-20 student trained with privileged training. The reward is computed by running each policy on uniformly sampled terrains.
    (\textbf{H-J})~Importance of the adaptive curriculum.
    }\label{fig:5}
\end{figure*}
\begin{figure*}
    \centering
    \includegraphics[width=\textwidth]{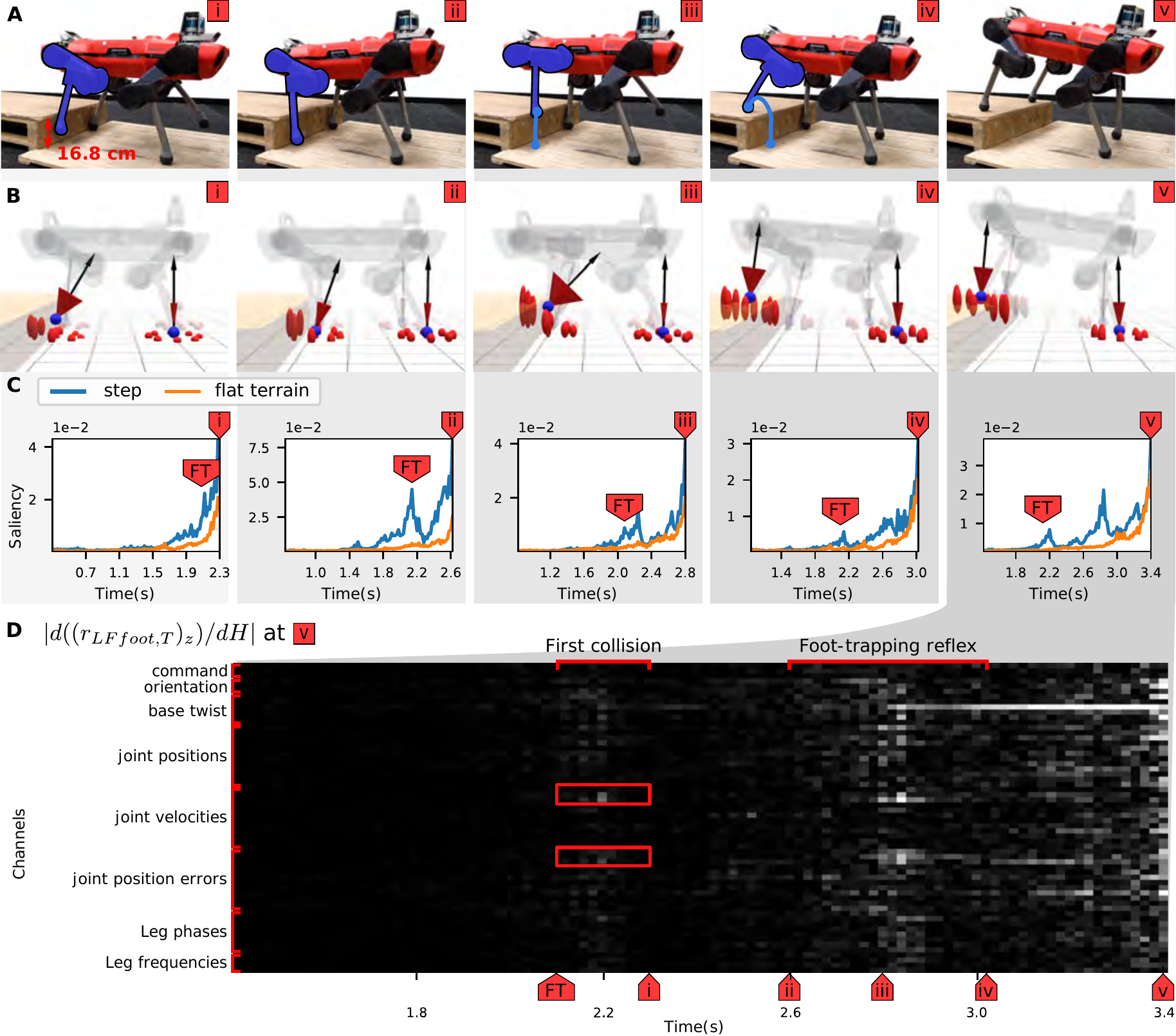}
    \caption{\textbf{Analysis of the emergent foot-trapping reflex.} FT denotes the first contact of the LF foot with the step (foot-trapping event).
    (\textbf{A}) The LF foot hits the step and then manifests higher foot clearance to overcome the step (ii-iv) in the following swing phase. 
    (\textbf{B}) Reconstructed terrain information from TCN embeddings.
     Red ellipsoid: estimated terrain shape around the foot. The center of the ellipsoid refers to the estimated terrain elevation and the vertical length represents uncertainty (standard deviation). Black arrow: terrain normal at the in-contact foot. Red cone: uncertainty of normal estimation. Blue spheres: estimated in-contact feet.
    (\textbf{C}) Input saliency at different moments. The peaks show that the TCN policy attends to the foot-trapping (FT) that happened around \unit[2.1]{s}. The orange curve (flat terrain) shows the saliency value computed on a flat terrain at similar gait phases.
    (\textbf{D}) Saliency map unrolled across input channels at \unit[3.4]{s}. Red boxes refer to joint measurements from the LF leg at the moment it collides with the step.}
    \label{fig:6}
\end{figure*}

\subsection*{Validation of the method}
We present ablation studies to justify each component of our approach: (1) using a sequence model for the student policy, (2) privileged training, and (3) adaptive terrain curriculum.

\subsubsection*{Memory in proprioceptive control}

We evaluate the importance of incorporating proprioceptive memory in the controller via the TCN architecture~\cite{bai2018empirical}.
Let TCN-$N$ denote a TCN with a receptive field of $N$ time steps. The network architectures we use are specified in detail in Table~S5.
We test controllers in diagnostic settings designed to focus on specific capabilities. Specifically, we test omnidirectional locomotion on sloped ground, traversal of a discrete step, and robustness to external disturbances (Fig.~\ref{fig:5}A).

Fig.~\ref{fig:5}B-D summarizes the importance of the memory length $N$. In these experiments, $N$ is varied from $1$ (corresponding to 20 ms of memory) to $100$ (2 s of proprioceptive memory). The latter is the default setting used in our deployed controller.

As shown in Fig.~\ref{fig:5}B, memory length doesn't have a strong effect in the uniform slope setting. Memory length does have a strong effect on the controller's ability to traverse a step (Fig.~\ref{fig:5}B-C). Controllers with longer memory are able to handle higher steps. As shown in Fig.~\ref{fig:5}C, the failure rate of limited-memory controllers is particularly high when the hind legs encounter the step.
Controllers with longer memory also adapt hind leg trajectories to ensure higher foot clearances.

Fig.~\ref{fig:5}D shows that controllers with longer memory are more robust to external disturbances.
We applied an external \unit[50]{N} force laterally to the base for \unit[5]{s} during a straight walk and evaluated the resulting deviation from the intended locomotion direction. The deviation of the TCN-100 controller was \unit[35.5]{\%} lower than that of TCN-1.

\subsubsection*{Privileged training}

We now assess the importance of privileged training. As a baseline, we train a TCN-20 policy directly, without the two-stage privileged training protocol. The policy is trained by TRPO~\cite{schulman2015trust} with the same reward and hyperparameters that we use for teacher training. This baseline is compared to the same TCN-20 architecture trained via privileged learning.

The results are summarized in Fig.~\ref{fig:5}E-G. Fig.~\ref{fig:5}E shows that the baseline fails the diagnostic tests: it is incapable of locomoting on a slope or traversing a step. 
Fig.~\ref{fig:5}F shows that the baseline does not reach comparable reward during training as the teacher MLP architecture with privileged information or the proprioceptive TCN-20 architecture (same as the baseline, no privileged information) trained via privileged learning.
Fig.~\ref{fig:5}G shows the mean episode length during training, which indicates that the baseline fails to learn to balance and locomote.

\subsubsection*{Adaptive terrain curriculum}
We now evaluate the effect of the adaptive terrain curriculum on teacher training.
Terrains used for training (specifically, hills, steps, and stairs) are shown in Fig.~\ref{fig:4}B.
As a baseline, we trained a teacher using randomly generated terrains that are uniformly sampled from $\mathcal{C}$ as specified in Table~S2.
The success rates on the testing terrains are significantly lower when trained without the adaptive curriculum, as shown in Fig.~\ref{fig:5}H.
Fig.~\ref{fig:5}I shows that a teacher trained without adaptive curriculum plateaus at a lower reward level. Throughout the training process, the mean episode length is shorter for the model being trained without adaptive curriculum (Fig.~\ref{fig:5}J).
This is because uniform sampling is more likely to draw terrains that cannot be successfully traversed by the policy being trained. On these terrains, the policy fails early and receives less training signal as a result. The adaptive curriculum modulates the difficulty of sampled terrains so as to maximize the didactic benefit of each episode. We provide an additional evaluation of the adaptive curriculum in supplementary section~S6.

\subsection*{Further analysis of emergent behavior}\label{supp:foot-trapping}
Here we provide further analysis on how the proprioceptive policy adapts to different situations.

To investigate how the proprioceptive policy perceives the environment, we trained a decoder network which reconstructs the privileged information $x_t\in X$ from the output of an intermediate layer of a trained TCN policy. $x_t$ consists of information that is not directly observable by the student policy such as contact states, terrain shape, and external disturbances. For classification of foot contact states, we employ a standard cross-entropy loss function. For regression of other states, we predict both mean $m_i$ and standard deviation $\sigma_i$ for each component and use a negative Gaussian log-likelihood loss to quantify the uncertainty encoded in the TCN representation~\cite{kendall2017uncertainties}:
\begin{equation}
    \mathcal{L} = \sum_{i \in \dim (X \setminus \text{contact states})} \frac{(m_i-m_i^{gt})^2}{2\sigma_i^2} + \log(\sigma_i) 
\end{equation} with added weight decay. The superscript $gt$ refers to the ground truth generated in simulation. Note that the parameters of the policy network are fixed during decoder training. Therefore, the decoder network is not used for policy training. It only provides insight into the information encoded by the TCN policy after training.

In Fig.~\ref{fig:6}, we provide snapshots of the foot-trapping reflex motion (Fig.~\ref{fig:6}A) and the reconstructed privileged information. In Fig.~\ref{fig:6}B we show the reconstructed terrain geometry and foot contact state. When the LF foot collides with the step, the estimated elevation in front of the front legs increases and its uncertainty grows (i+ii). The estimated elevations and normal vectors adapt to the step during the foot-trapping reflex (iii+iv). After the successful step-up, the terrain uncertainty remains elevated (v), indicating an anticipation of generally rough terrain. Additionally, the decoder network can detect foot contacts with horizontal and vertical surfaces while successfully identifying frontal collision as such, as indicated by the estimated terrain normal vector (i + iii). The ability to reconstruct explicit environmental information from the encoding of the proprioceptive history is a strong indicator that the TCN policy learns to build an internal representation of the environment and uses it for decision making. We provide more examples of the reconstructed privileged information in supplementary section~S7.

We then analyse how the proprioceptive policy leverages past observations. We compute the saliency map of the input $H \in \mathbb{R}^{60 \times N}$, and visualize the sensitivity of the policy to each element of the input while overcoming the step~\cite{simonyan2013deep}.
Each column of $H$ is a proprioceptive measurement $h \in \mathbb{R}^{60}$, and we stack $N$ measurements (history length = $\unit[0.02]{s}\times N$).
We define the saliency value for the $i$-th measurement ($i \in [0, N]$) as

\begin{equation}
    M_i = \sum_{j \in channels}(\lvert d((r_{f,T})_z)/dH_{i,j} \rvert) \in \mathbb{R},
\end{equation}

where $(r_{f,T})_z$ refers to the height command for the foot $f$.
We computed the value for  $(r_{f,T})_z$ because we are interested in the change in foot clearance.
$M_i$ can be interpreted as the sensitivity of the output to the $i$th measurement.
As we use 1D convolution over time, the output is in $\mathbb{R}^{N}$, i.e., each row of $H$ is regarded as a channel.

In Fig.~\ref{fig:6}C we can see that the saliency value at the foot-trapping event (FT) is kept high while stepping up.
The policy has direct access to the measurements at the moment of foot-trapping, and leverages this in the following swing phase.
This is highlighted by the red boxes in Fig.~\ref{fig:6}D.
The policy attends to the LF leg joint states measured at the foot-trapping event.

\section{Acknowledgments}
\textbf{Funding}
The project was funded, in part, by the Intel Network on Intelligent Systems, the Swiss National Science Foundation (SNF) through the National Centre of Competence in Research Robotics, the European Research Council (ERC) under the European Union’s Horizon 2020 research and innovation programme grant agreement No 852044 and No 780883. The work has been conducted as part of ANYmal Research, a community to advance legged robotics.
\textbf{Author contribution}
J.L formulated the main idea of the training and control methods, implemented the controller, set up the simulation, and trained control policies.
J.L performed the indoor experiments.
J.H contributed in setting up the simulation.
J.L and L.W performed outdoor experiments together.
J.L, J.H, L.W, M.H, and V.K refined ideas, contributed in the experiment design and analyzed the data.
\textbf{Conflict of interest} The authors declare that they have no competing interests.
\textbf{Data and materials availability} All (other) data needed to evaluate the conclusions in the paper are present in the paper or the Supplementary Materials. Other materials can be found at \href{https://github.com/leggedrobotics/learning_quadrupedal_locomotion_over_challenging_terrain_supplementary}{\url{https://github.com/leggedrobotics/learning_quadrupedal_locomotion_over_challenging_terrain_supplementary}}.

\section*{Supplementary materials}
\makebox[1.8cm][l]{Section S1.} Nomenclature \\
\makebox[1.8cm][l]{Section S2.} Implementation details\\
\makebox[1.8cm][l]{Section S3.} Foot trajectory generator\\
\makebox[1.8cm][l]{Section S4.} Reward function for teacher policy training \\
\makebox[1.8cm][l]{Section S5.} Parameterized terrains\\
\makebox[1.8cm][l]{Section S6.} Qualitative evaluation of the adaptive terrain curriculum\\
\makebox[1.8cm][l]{Section S7.} Reconstruction of the privileged information \\
\makebox[1.8cm][l]{}in different situations\\
\makebox[1.8cm][l]{Section S8.} Recurrent neural network student policy \\
\makebox[1.8cm][l]{Section S9.} Ablation of the latent representation loss for student\\
\makebox[1.8cm][l]{}training\\
\makebox[1.8cm][l]{Algorithm S1.} Teacher training with automatic terrain curriculum\\
\makebox[1.8cm][l]{Figure S1.} Illustration  of  the adaptive curriculum.\\
\makebox[1.8cm][l]{Figure S2.} Reconstructed privileged information in different\\
\makebox[1.8cm][l]{}situations.\\
\makebox[1.8cm][l]{Figure S3.} Comparison of neural network architectures for the\\
\makebox[1.8cm][l]{}proprioceptive controller \\
\makebox[1.8cm][l]{Table S1.} Computation time for training \\
\makebox[1.8cm][l]{Table S2.} Parameter spaces $\mathcal{C}$ for simulated terrains\\
\makebox[1.8cm][l]{Table S3.} Hyperparameters for automatic terrain curriculum\\
\makebox[1.8cm][l]{Table S4.} State representation for proprioceptive controller and\\
\makebox[1.8cm][l]{}the privileged information \\
\makebox[1.8cm][l]{Table S5.} Neural network architectures \\
\makebox[1.8cm][l]{Table S6.} Network parameter settings and the training time for\\
\makebox[1.8cm][l]{}student policies\\
\makebox[1.8cm][l]{Table S7.} Hyperparameters for teacher policy training \\
\makebox[1.8cm][l]{Table S8.} Hyperparameters for student policy training \\
\makebox[1.8cm][l]{Table S9.} Hyperparameters for decoder training\\
\href{https://youtu.be/txjqn8h6pjU}{\makebox[1.8cm][l]{Movie S1.} Deployment in a forest} \\
\href{https://youtu.be/Xnn4sVSpSh0}{\makebox[1.8cm][l]{Movie S2.} Locomotion over unstable debris} \\
\href{https://youtu.be/tPixnjLbTvE}{\makebox[1.8cm][l]{Movie S3.} Step experiment}   \\
\href{https://youtu.be/3Nr47MXCFO0}{\makebox[1.8cm][l]{Movie S4.} Payload experiment}  \\
\href{https://youtu.be/aMPwB3t4idU}{\makebox[1.8cm][l]{Movie S5.} Foot slippage experiment}

\clearpage
\newpage

\setcounter{table}{0}
\makeatletter 
\renewcommand{\thetable}{S\@arabic\c@table}
\makeatother

\setcounter{figure}{0}
\makeatletter 
\renewcommand{\thefigure}{S\@arabic\c@figure}
\makeatother

\setcounter{algorithm}{0}
\makeatletter 
\renewcommand{\thealgorithm}{S\@arabic\c@algorithm}
\makeatother

\section*{Supplementary materials}
\subsection*{S1. Nomenclature}
\makebox[1.2cm]{$\hat{(\cdot)}$} normalized vector\\
\makebox[1.2cm]{$\dot{(\cdot)}$} first derivative\\
\makebox[1.2cm]{$\bar{(\cdot)}$} teacher's quantity\\
\makebox[1.2cm]{${(\cdot)}_T$} target quantity\\
\makebox[1.2cm]{$^C_{AB}v$} linear velocity of $B$ frame with respect to $A$ frame \\
\makebox[1.2cm]{}\. expressed in $C$ frame\\
\makebox[1.2cm]{$c_T$} terrain parameter vector\\
\makebox[1.2cm]{$\omega$} angular velocity\\
\makebox[1.2cm]{$\tau$} joint torque\\
\makebox[1.2cm]{$\theta$} joint angle\\
\makebox[1.2cm]{$\psi$} yaw angle\\
\makebox[1.2cm]{$\phi$} leg phase\\
\makebox[1.2cm]{$f$} leg frequency\\
\makebox[1.2cm]{$r_f$} linear position of a foot\\
\makebox[1.2cm]{$e_g$} gravity vector\\
\makebox[1.2cm]{$H$} horizontal frame\\
\makebox[1.2cm]{$g_i$} gap function of the $i$-th possible contact pair\\
\makebox[1.2cm]{$I_{c}$} index set of all contacts\\
\makebox[1.2cm]{$I_{c,body}$} index set of body contacts\\
\makebox[1.2cm]{$I_{c,foot}$} index set of foot contacts\\
\makebox[1.2cm]{$I_{swing}$} index set of swing legs \\
\makebox[1.2cm]{$\lvert \cdot \rvert$} cardinality of a set or $l_1$ norm\\
\makebox[1.2cm]{$\lvert\lvert \cdot \rvert\rvert$} $l_2$ norm\\

\subsection*{S2. Implementation details}\label{impl_details}
The RaiSim simulator~\cite{hwangbo2018per} is used for rigid-body and contact dynamics simulation.
The actuator networks~\cite{hwangbo2019learning} are trained for each robot to simulate Series Elastic Actuators (SEA)~\cite{pratt1995series} at the joint.
The input to the actuator model is a 6-dimensional real-valued vector consisting of the joint position error and velocity at current time step $t$ and two past states corresponding to $t - \unit[0.01]{s}$ and $t - \unit[0.02]{s}$.
The feature selection is done as in~\cite{hwangbo2019learning}.
%and proven to be sufficient to learn the dynamics of SEAs in the operating range of our tasks.

As several studies have shown that randomization of dynamic properties improves the robustness of the policy~\cite{hwangbo2019learning,tan2018sim}, we also randomized several physical quantities, and the teacher policy has access to these values during training.
We applied disturbances, randomized friction coefficients between the feet and the terrain, and additive noise to the observations during training.
% we did not apply dynamic randomization as we did not observe any difference.

The training process for the teacher policy is depicted in Algorithm~S1. Hyperparameters are given in Table~S3. In our implementation of the terrain curriculum, we update the curriculum every $N_{evaluate}$ policy iterations to reduce variance. We assume that within $N_{evaluate}$ iterations, the performance of the policy is similar. With the slower update rate, the measurement probability of Eq.6 becomes
\begin{equation}
    \Pr(y_j^k|c_{T,j}^k) 
    \approx
    \sum_{}^{N_{evaluate}}
     \sum_{}^{N_{traj}} 
     \frac{\mathbbm{1} (Tr(c_{T,j}^k,\pi) \in [0.5, 0.9])}{ N_{traj} N_{evaluate}}.
\label{measurement_model3}\end{equation} Additionally, we leverage replay memory to prevent degeneration of the particle filter and to avoid catastrophic forgetting.

The controller is implemented with a state machine to switch between the ``standing still'' state and the locomotion state. We set the base frequency $f_0$ to zero when the zero command is given for \unit[0.5]{s}, which stops FTGs, and the robot stands still on the terrain.
$f_0$ is set to \unit[1.25]{Hz} when the direction command is given or the linear velocity of the base exceeds \unit[0.3]{m/s} for the disturbance rejection. The state machine is included in the training environment.

During the deployment, the base velocity and orientation are estimated by the state estimator that relies on inertial measurements and leg kinematics~\cite{bloesch2013state}.

The neural network policy runs at 400 Hz on an onboard CPU (Intel i7-5600U, 2.6 -- 3.2GHz, dual-core 64-bit) integrated into the robot. The Tensorflow C++ API is used for onboard inference.

\begin{algorithm}
\caption{Teacher training with automatic terrain curriculum}
\begin{algorithmic}[1]
\State Initialize a replay memory, Sample $N_{particle}$ $c_{T,0}$s uniformly from $\mathcal{C}$ (Table~S2), $i,j  = 0$.
\Repeat 
\For{$0 \leq k \leq N_{evaluate}$}
\For{$0 \leq l \leq N_{particle}$} 
\For{$0 \leq m \leq N_{traj}$} 
% \Comment{This loop runs in parallel}
\State Generate terrain using  $c_{T,j}^l$
\State Initialize robot at random position
\State Run policy $\pi_{i}$
\State Compute traverability label  for each
\par \hskip\algorithmicindent \hskip\algorithmicindent \hskip\algorithmicindent state transition (Eq.~2)
\State Save the scores and the trajectory
\EndFor
\EndFor
\State Update policy using TRPO~\cite{schulman2015trust}
\State $i = i + 1$
\EndFor

\For{$0 \leq l \leq N_{particle}$} 
\State Compute measurement probability for each
\par \hskip\algorithmicindent 
parameter $c_{T,j}^l$s~(Eq.~9)
\EndFor
\For{$0 \leq l \leq N_{particle}$} 
\State Update weights $w_{j} = \frac{P(y_i^l|c_{T,j}^l)}{\sum_{m} P(y_i^m|c_{T,j}^m)}$
\EndFor
\State Resample $N_{particle}$ parameters
\State Append $c_{T,j}$s to the replay memory
\For{$0 \leq l \leq N_{particle}$} 
\State by $p_{replay}$ probability, sample from replay memory
\State by $p_{transition}$ probability, move $c^l_{T,j}$ to an adjacent value 
\par \hskip\algorithmicindent
in $\mathcal{C}$.

\EndFor
\State $j = j + 1$
% \State $c_{T,j} = c_{T,j-1} + \epsilon$ for all particles
\Until{Convergence}
\end{algorithmic}
\label{algo:TPF}
\end{algorithm}

\subsection*{S3. Foot trajectory generator}

The foot trajectory is defined as
\begin{equation}
F(\phi_i) = 
\begin{cases}
(h (-2k^3+3k^2) - 0.5) ^{H_i}z  &  k \in [0, 1] \\
(h (2k^3-9k^2+12k-4) - 0.5)  ^{H_i}z &  k \in [1, 2] \\
- 0.5 ^{H_i}z & \text{otherwise},
\end{cases}{}
% \quad \text{where} \quad k = 2\phi / \pi, \, \phi_i = f_i t\pmod{2 \pi} - \pi
\end{equation}
where $k = 2 (\phi_i - \pi) / \pi$ and $h$ is a parameter for the maximum foot height.
Each segment during the swing phase ($k \in [0, 2)$) is a cubic Hermite spline connecting the highest and lowest points with a zero first derivative at the connecting points.
Other periodic functions such as $ h_i \sin(\phi_i)$ can be used for the FTG.
With a set of reasonably tuned $f_0$, $h$ and $\phi_{i,0}$, a quadruped can stably step in place.
In our setting, $f_0$ = 1.25, $h$ = \unit[0.2]{m}, and $\phi_{i,0}$ are sampled from $U(0, 2\pi)$.

\subsection*{S4. Reward function for teacher policy training}

The reward function is defined as
$0.05r_{lv} + 0.05r_{av} + 0.04r_b + 0.01r_{fc} + 0.02r_{bc} + 0.025 r_s+ 2 \cdot 10^{-5} r_{\tau}.$
% $2.5r_{lv} + 2.5r_{av} + 2.0r_b + 0.5r_{fc}+1.0r_{bc}+1.2r_s+0.001r_{tau}.$
The individual terms are defined as follows.

\begin{itemize}
    \item Linear Velocity Reward ($r_{lv}$): This term maximizes the  $v_{pr} = (^B_{IB}v)_{xy} \cdot (^{B}_{IB} \hat{v}_{T})_{xy}$, which is the base linear velocity projected onto the command direction.
    \begin{equation}
    r_{lv} \coloneqq
        \begin{cases}
            \exp{(-2.0 (v_{pr} - 0.6)^2)} & v_{pr} < 0.6 \\
            1.0  & v_{pr} \geq 0.6 \\
           % 0.0  & \lvert \lvert  (^B_{IB}\hat{v})_{xy} \rvert \rvert = 0.0 \\
            0.0  & \text{zero command}\\
        \end{cases}.
    \end{equation}
    The velocity threshold is defined as \unit[0.6]{m/s} which is the maximum speed reachable on the flat terrain with the existing controller~\cite{bellicoso2018dynamic}.
   
   \item Angular Velocity Reward ($r_{av}$): We motivate the agent to turn as fast as possible along the base $z$-axis when $(^{B}_{IB}\hat{\omega}_{T})_{z}$ is nonzero. It is defined as
    \begin{equation}
    r_{av} \coloneqq 
        \begin{cases}
            \exp{(- 1.5 ( \omega_{pr} - 0.6)^2)} & \omega_{pr} < 0.6 \\
            1.0  & \omega_{pr} \geq 0.6 \\
        \end{cases},
    \end{equation}
     where $\omega_{pr} = (^B_{IB} \omega)_{z} \cdot (^{B}_{IB}\hat{\omega}_{T})_{z}$.
    
    \item Base Motion Reward ($r_{b}$): 
    This term penalizes the velocity orthogonal to the target direction and the roll and pitch rates such that the base is stable during the locomotion.
    \begin{equation}
      r_{b} \coloneqq \exp(- 1.5 v_{o}^2) + \exp( - 1.5 \lvert\lvert(^B_{IB} \omega)_{xy} \rvert\rvert^2)
    \end{equation}
   where $v_{o} = \lvert\lvert (^B_{IB} v)_{xy} - v_{pr}\cdot (^{B}_{IB} \hat{v}_{T})_{xy} \rvert\rvert$.
    When the stop command is given, $v_{o}$ is replaced by $\lvert\lvert ^B_{IB} v\rvert\rvert$.
    
    \item Foot Clearance Reward ($r_{fc}$):
    When a leg is in swing phase, i.e., $\phi_i \in [\pi, 2\pi)$, the robot should lift the corresponding foot higher than the surroundings to avoid collision.
    We first define the set of such collision-free feet as $\mathcal{F}_{clear} = \{ i : r_{f,i} > max(H_{scan, i}) , i \in I_{swing} \}$,
    where $H_{scan,i}$ is the set of scanned heights around the $i$-th foot.
    Then the clearance cost is defined as
    \begin{equation}
        r_{fc} \coloneqq \sum_{i \in I_{swing}} 
        (\mathbbm{1}_{\mathcal{F}_{clear}}(i) / \lvert I_{swing} \rvert )
        \in [0.0, 1.0].
    \end{equation}
    
    \item Body Collision Reward ($r_{bc}$): We want to penalize undesirable contact between the robot's body parts and the terrain to avoid hardware damage. 
    \begin{equation}
        r_{bc} \coloneqq -\lvert I_{c, body} \backslash I_{c, foot} \rvert.
    \end{equation}
   
    \item Target Smoothness Reward ($r_{s}$): 
    The magnitude of the second order finite difference derivatives of the target foot positions are penalized such that the generated foot trajectories become smoother.
     \begin{equation}
        r_{s} \coloneqq - \lvert\lvert  
        (r_{f,d})_{t} - 2(r_{f,d})_{t-1} + (r_{f,d})_{t-2}
        \rvert\rvert .
    \end{equation}
    
    \item Torque Reward ($r_{\tau}$): We penalize the joint torques to prevent damaging joint actuators during the deployment and to reduce energy consumption ($\tau \propto \text{electric current}$).
    \begin{equation}
          r_{\tau} \coloneqq - \textstyle \sum_{i\in joints} \lvert \tau_i \rvert.
    \end{equation}

\end{itemize}

\subsection*{S5. Parameterized terrains}
It is important to generate training environments that can pose representative challenges such as foot slippage and foot-trapping.
To efficiently synthesize random terrains, we use procedural generation techniques~\cite{smelik2009survey}.
This method allows us to generate a large number of different terrains by changing a set of terrain parameters $c_T \in \mathcal{C}$.
In the following, we describe the three terrain generators used in this work.
See Fig.~4B for a visualization of the terrains and Table~S2 for the definition of parameter spaces $\mathcal{C}$.

\begin{itemize}
  \item The \textit{Hills} terrain is based on Perlin noise~\cite{lagae2010survey}.
  The terrain is generated via three parameters: roughness, frequency of the Perlin noise, and amplitude of the Perlin noise.  
%   $c_{T} \coloneqq \langle \text{roughness, frequency, amplitude} \rangle$
  The height of each element of the output height map $hm$ is defined as $hm[i,j] \coloneqq Perlin (c_{T,2}, c_{T,3})[i,j] + U(-c_{T,1}, c_{T,1})$.
  A policy experiences smooth slopes and foot slippage on this terrain during training.
  
  \item The \textit{Steps} terrain consists of square steps of random height. For every $c_{T,1}$ by $c_{T,1}$ blocks, the height is sampled from $U(0, c_{T,2})$.
  A policy experiences discrete elevation changes and foot-trapping on this terrain.
  \item The \textit{Stairs} terrain is a staircase with fixed width and height. The robot is initialized at the flat segment in the middle of the staircase  (see Fig.~\ref{fig:4}B).
\end{itemize}
The ranges are defined considering the kinematics of the robot, e.g., a step height should be lower than leg length.
During training, the terrain is regenerated every episode with a different random seed.

\subsection*{S6. Qualitative evaluation of the adaptive terrain curriculum}
The behavior of adaptive curriculum is illustrated in Fig.~S1.
Fig.~S1A-C focuses on the \textit{Hills} terrain type. There are three parameters for this terrain: roughness, frequency, and amplitude. 
The relationship between traversability (Eq.~3) and desirability (Eq.~4) is illustrated in Fig.~S1A-B.
Undesirable terrains are either too easy or too difficult, as shown in the leftmost and rightmost panels of Fig.~S1A.
Fig.~S1B-C shows that the particle filter fits the latent distribution of desirable terrains, which has a bow shape in the frequency-amplitude marginal (middle).
Fig.~S1D focuses on the \textit{Stairs} terrain type and shows the evolution of terrain parameters during training.
The particle filter rejects parameters that represent short and steep steps (upper-left area). The curriculum initially focuses on wide and shallow steps (middle panels, particularly Iter. 50-60), and then broadens the distribution to include narrower steps (rightmost panels).

\subsection*{S7. Reconstruction of the privileged information in different situations}
In Fig.~S2, we provide the decoded privileged information in different situations. Fig.~S2A shows the estimated friction coefficient between the feet and the terrain when traversing a wet, slippery whiteboard, as shown in \href{https://youtu.be/aMPwB3t4idU}{Movie~S5}. The estimate decreases as soon as the first foot starts slipping (i), remains low throughout the traversal (ii) and increases about \unit[2]{s} after the robot returns to normal ground (iii). The external disturbance and terrain information can also be reconstructed from the TCN embedding.
As shown in Fig.~S2B, the decoder detects downward external force when an unknown \unit[10]{kg} payload is applied.
While traversing dense vegetation as shown in Fig.~S2C, it detects a force opposite the motion direction, which makes the policy to counteract and push through the vegetation.
The uncertainty of the elevation estimates are notably high in the natural terrains shown in Fig.~S2C and Fig.~S2D, which indicates that the TCN policy encodes the roughness of the terrain.

\subsection*{S8. Recurrent neural network student policy}

We use the TCN architecture for the proprioceptive policy~\cite{bai2018empirical}. For comparison, we also evaluated a recurrent network with gated recurrent units~(GRU)~\cite{chung2014empirical}. The architectures are specified in Tables~S5 and~S6.
The loss function for training a GRU student policy is defined as
\begin{equation}
\mathcal{L} \coloneqq (\bar{a_t}(o_t, x_t) - a_t(o_t))^2 +  (\bar{l_t}(o_t, x_t) - l_t(o_t))^2. 
\end{equation}
To improve the performance and computational efficiency of the training, we have implemented Truncated Backpropagation Though Time (Truncated BPTT)~\cite{williams1990efficient}.

Performance on the diagnostic settings presented in Fig.~5A is given in Fig.~S3.
Overall, the performance of the GRU-based controller is between that of TCN-20 and TCN-100.
The performance is comparable to TCN-100 in the slope setting, but the GRU-based controller fails to achieve the performance of TCN-100 in step experiments.

The chief advantage of the TCN is in training efficiency. The training time for the TCN is much faster in comparison to the GRU. The computation times are reported in Table~\ref{tab:times}. 

\subsection*{S9. Ablation of the latent representation loss for student training}

We examine the effect of the second term in the loss function for student policy training presented in Eq.~1, which is a squared error loss for the latent vector $l_t$.
As a baseline, we train a student policy using the following loss function:
\begin{equation}
\mathcal{L} \coloneqq (\bar{a_t}(o_t, x_t) - a_t(o_t, H))^2,
\label{eq:student2}\end{equation}
which simply imitates the output of the teacher.

The result is reported in Fig.~S3 as `TCN-100 naive IL'.
The performance is comparable in the uniform slope setting and under external disturbances. On the other hand, the ablated version has lower success rates on steps.

\begin{figure*}
    \centering
    \includegraphics[width=5.0in]{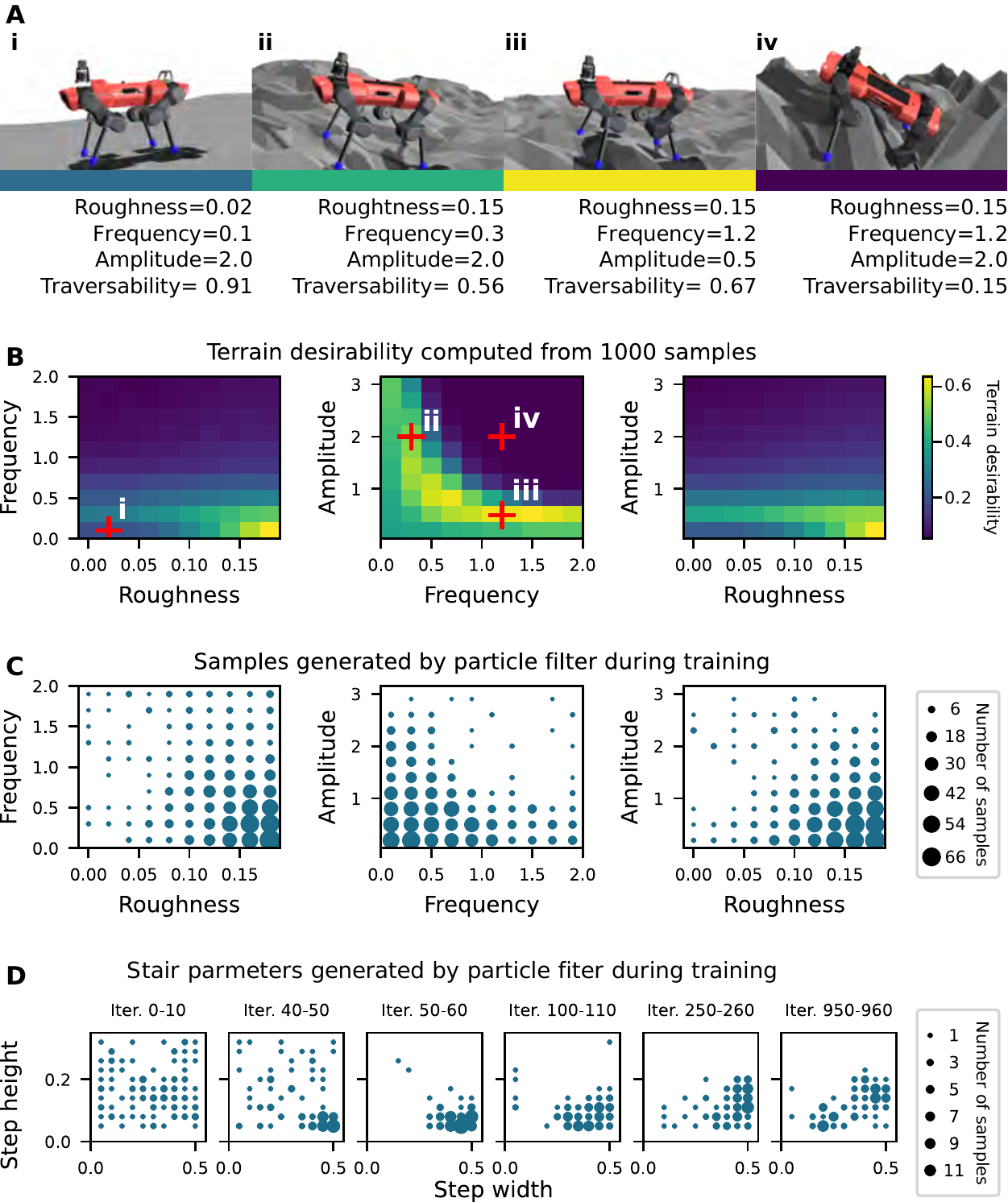}
    \caption{\textbf{Illustration of the adaptive curriculum.}
    (\textbf{A}) Examples of \textit{Hills} terrains. The color bar indicates desirability; dark blue represents low desirability.
    (\textbf{B}) Terrain desirability estimated from 1000 trajectories generated by a fully trained teacher policy. The red crosses correspond to the examples presented in A.
    (\textbf{C}) The distribution of terrain profiles sampled by the particle filter during the last 100 iterations of teacher training.
    (\textbf{D}) Evolution of \textit{Stairs} terrain parameters during training.}\label{fig:tpf}
\end{figure*}

\begin{figure*}
    \centering
    \includegraphics[width=5.0 in]{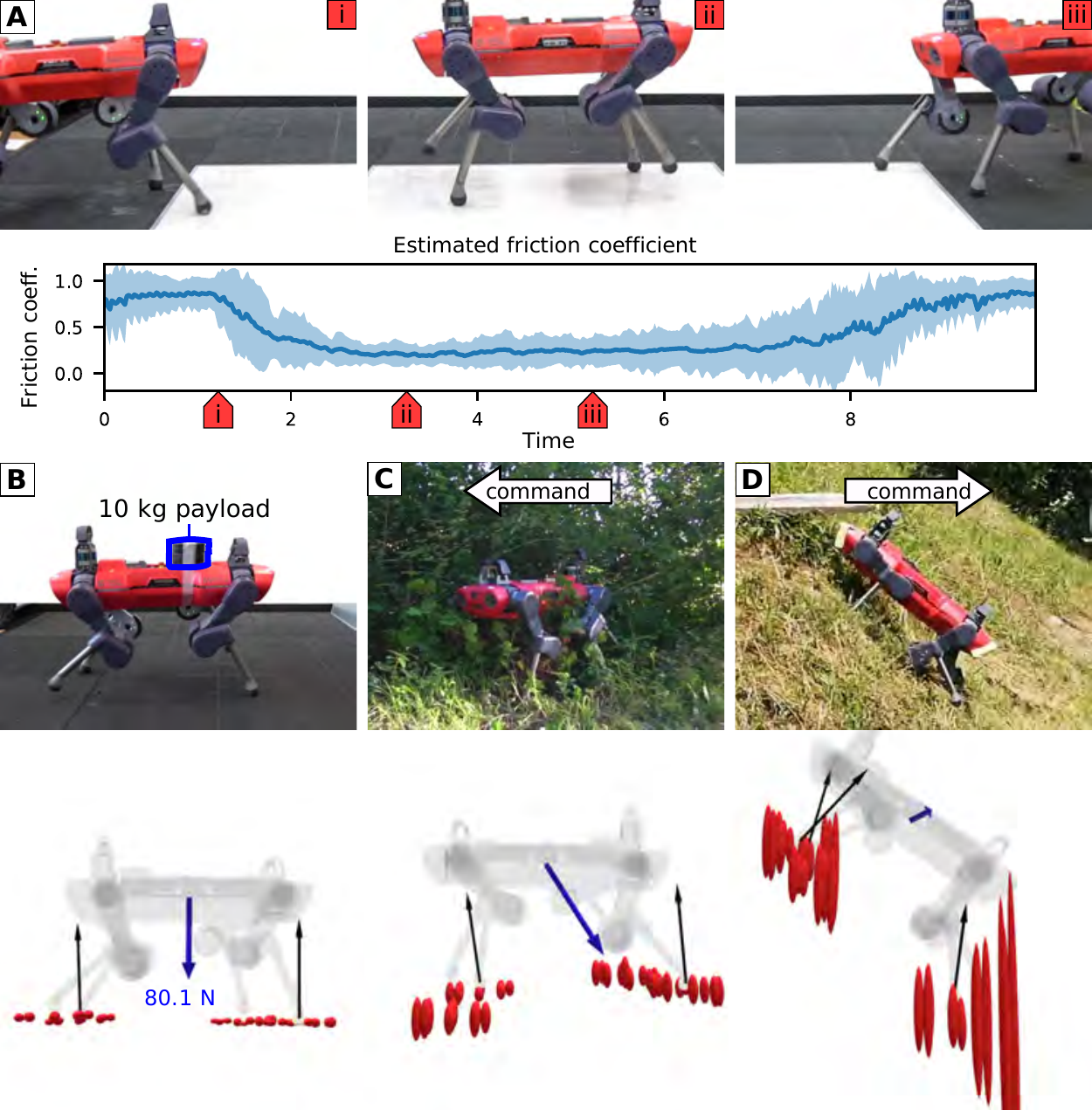}
    \caption{\textbf{Reconstructed privileged information in different situations.}
    (\textbf{A}) Estimated friction coefficient between the feet and the terrain while traversing a wet whiteboard. The shaded area denotes \unit[95]{\%} confidence interval. (\textbf{B-D}) Reconstruction of the  external disturbance and terrain information in different scenarios. Blue arrow: estimated external force applied to the torso. Red ellipsoid: estimated terrain shape around the foot. The center of the ellipsoid refers to the estimated terrain elevation and the vertical length represents uncertainty (1 standard deviation). For each foot, 8 ellipsoids are symmetrically placed along a circle with \unit[10]{cm} radius. Black arrow: terrain normal at the in-contact foot.
    }\label{fig:decoding}
\end{figure*}

\begin{figure*}
    \centering
    \includegraphics[width=3.5 in]{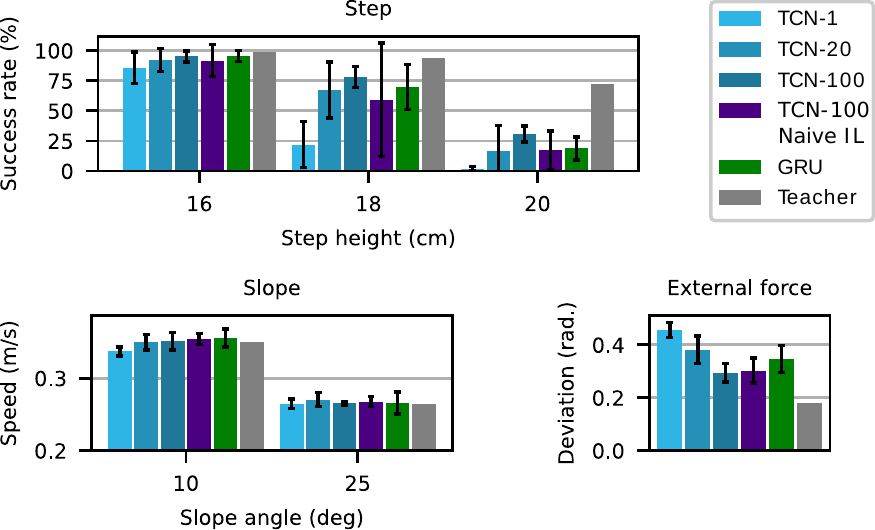}
    \caption{\textbf{Comparison of neural network architectures for the proprioceptive controller.} We trained each model 5 times using different random seeds. The error bars denote 95\% confidence intervals. `TCN-100 naive IL' denotes the TCN-100 network trained using a naive imitation learning method without the latent representation loss (Eq.~19).}
    \label{fig:supp_student_comparison}
\end{figure*}

\clearpage
\newpage

\begin{table*}
\centering
 \begin{tabular}{|c|c|c|c|}
\hline 
 Name & Time \\ \hline
Teacher policy training & $\approx$ \unit[12]{hrs}  \\ 
Student policy training & $\approx$ \unit[4]{hrs} \\ 
Adaptive terrain curriculum & \unit[2.9]{s}   \\ 
\hline
\end{tabular}
\caption{\textbf{Computation time for training}. The TCN-100 architecture is used for the student policy. The training is conducted on a desktop machine with i7-8700K CPU and a Geforce RTX 2080 GPU.} \label{tab:times}
\end{table*}

\begin{table*}
\centering
 \begin{tabular}{|c|c|c|c|c|}
\hline
Terrain & grid size & friction coefficient& parameters ($c_{T}$) & range \\ \hline
\multirow{3}{*}{\textit{Hills}} & \multirow{3}{*}{\unit[0.2]{m}}  & \multirow{3}{*}{ $\mathcal{N}(0.7, 0.2)$ }  & roughness ($m$) & $[0.0, 0.05]$ \\ 
 & & & frequency & $[0.2, 1.0]$ \\ 
 & & & amplitude ($m$) & $[0.2, 3.0]$ \\ \hline
 \multirow{3}{*}{\textit{Slippery Hills}} & \multirow{3}{*}{\unit[0.2]{m}}  & \multirow{3}{*}{$\mathcal{N}(0.3, 0.1)$ }  & roughness ($m$) & $[0.0, 0.05]$ \\ 
 & & & frequency & $[0.2, 1.0]$ \\ 
 & & & amplitude ($m$) & $[0.2, 3.0]$ \\ \hline
 \multirow{2}{*}{\textit{Steps}}  & \multirow{2}{*}{\unit[0.02]{m}}  & \multirow{2}{*}{$\mathcal{N}(0.7, 0.2)$}  & step width ($m$) & $[0.1, 0.5]$ \\ 
& & & step height ($m$) & $[0.05, 0.3]$ \\ \hline
  \multirow{2}{*}{\textit{Stairs}}  & \multirow{2}{*}{\unit[0.02]{m}}  & \multirow{2}{*}{$\mathcal{N}(0.7, 0.2)$}  & step width ($m$) & $[0.1, 0.5]$ \\
& & & step height ($m$) & $[0.02, 0.2]$ \\ \hline
\end{tabular}
\caption{\textbf{Parameter spaces $\mathcal{C}$ for simulated terrains.} $\mathcal{N}(m, d)$ denotes that the value is sampled from the Gaussian distribution of mean $m$ and stardard deviation $d$. The friction coefficient is clipped to be above 0.1.} \label{tab:terrainparams}
\end{table*}

\begin{table*}
    \centering
    \begin{tabular}{|c|c|} \hline
        Parameter & value \\ \hline
        Number of particles ($N_{particle}$) & 10 per terrain type \\ \hline
        Transition probability ($p_{transition}$) & 0.8 \\ \hline
        Trajectories per particle ($N_{traj}$) & 6 \\ \hline
        Update rate of the terrain parameters ($N_{evaluate}$) & 10 \\ \hline
        Probability of sampling from replay memory ($P_{replay}$) & 0.05 \\ \hline
    \end{tabular}
    \caption{\textbf{Hyperparameters for automatic terrain curriculum.}}
    \label{tab:pfparams}
\end{table*}

\begin{table*}
\centering
 \begin{tabular}{|l|c|c|c|c|}
 \hline
Data & dimension & $x_{t}$ &$o_{t}$ &$h_{t}$ \\
\hline
                Desired direction ($(^{B}_{IB} \hat{v}_{d})_{xy}$) & 2 &  & \checkmark & \checkmark \\
                Desired turning direction ($(^{B}_{IB}\hat{\omega}_{d})_{z}$) & 1 &  & \checkmark & \checkmark \\
                Gravity vector ($e_g$) & 3 &  & \checkmark & \checkmark \\
                Base angular velocity ($^{B}_{IB} \omega$) & 3 &  & \checkmark & \checkmark \\
                Base linear velocity ($^{B}_{IB} v$) & 3 &  & \checkmark & \checkmark \\
                Joint position/velocity ($\theta_{i}$, $\dot{\theta_{i}}$) & 24 &  & \checkmark & \checkmark \\
                FTG phases ($\sin(\phi_i)$, $\cos(\phi_i)$) & 8 &  & \checkmark & \checkmark \\
                FTG frequencies ($\dot{\phi}_i$) & 4 &  & \checkmark & \checkmark \\
                Base frequency ($f_o$) & 1 & &\checkmark &  \\
                Joint position error history & 24 &  &  \checkmark  & \\
                Joint velocity history & 24 &  &  \checkmark  & \\
                Foot target history ($(r_{f,d})_{t-1, t-2}$) & 24 &  & \checkmark  &  \\
 \hline                 
                Terrain normal at each foot & 12 & \checkmark & & \\
                Height scan around each foot & 36 & \checkmark & & \\
                Foot contact forces & 4 & \checkmark & & \\
                Foot contact states & 4 & \checkmark & & \\
                Thigh contact states & 4 & \checkmark & & \\
                Shank contact states & 4 & \checkmark & & \\
                Foot-ground friction coefficients & 4 & \checkmark & & \\
                External force applied to the base & 3 & \checkmark & & \\
 \hline
 \end{tabular}
  \caption{\textbf{State representation for proprioceptive controller (top) and the privileged information (bottom).}} \label{tab:inputs}
\end{table*}

\begin{table*}
\centering
 \begin{tabular}{|c|c|c|c|c|c|c|c|}
\hline
Layer & \multicolumn{2}{|c|}{Teacher} &  \multicolumn{2}{|c|}{TCN-N Student} & \multicolumn{2}{|c|}{GRU Student}  & {Decoder} \\ \hline
input & $o_t$ & $x_t$ & $o_t$ & $h$ (60$\times$N) & $o_t$ & $o_t$ & {$\langle o_t, l_t \rangle$} \\ \hline
1 & id & tanh(72) & id & 1D conv dilation 1 & id & GRU(68)  & {relu(196)} \\ \hline
2 & id & tanh(64) & id & 1D conv stride 2 & \multicolumn{2}{|c|}{concatenate} & {Output} \\ \hline
3 & \multicolumn{2}{|c|}{concatenate }& id & 1D conv dilation 2 &  \multicolumn{2}{|c|}{tanh(256)*}  & {-} \\ \hline
4 &  \multicolumn{2}{|c|}{tanh(256)*}  &  id &  1D conv stride 2 &  \multicolumn{2}{|c|}{tanh(128)*} & {-}\\ \hline
5 & \multicolumn{2}{|c|}{tanh(128)*} & id &  1D conv dilation 4  &\multicolumn{2}{|c|}{tanh(64)*} & {-}\\ \hline
6 & \multicolumn{2}{|c|}{tanh(64)*} & id &  1D conv stride 2 & \multicolumn{2}{|c|}{Output*} & {-} \\ \hline
7 &  \multicolumn{2}{|c|}{Output*} & id & tanh(64) & \multicolumn{2}{|c|}{-} & {-} \\ \hline
8 &  \multicolumn{2}{|c|}{-} & \multicolumn{2}{|c|}{concatenate} & \multicolumn{2}{|c|}{-} & {-} \\ \hline
9 &  \multicolumn{2}{|c|}{-} & \multicolumn{2}{|c|}{tanh(256)*} & \multicolumn{2}{|c|}{-} & {-} \\ \hline
10 &  \multicolumn{2}{|c|}{-} & \multicolumn{2}{|c|}{tanh(128)*} & \multicolumn{2}{|c|}{-} & {-} \\ \hline
11 &  \multicolumn{2}{|c|}{-} & \multicolumn{2}{|c|}{tanh(64)*} & \multicolumn{2}{|c|}{-} & {-} \\ \hline
12 &  \multicolumn{2}{|c|}{-} & \multicolumn{2}{|c|}{Output*} & \multicolumn{2}{|c|}{-} & {-} \\ \hline
 \end{tabular}
 \caption{\textbf{Neural network architectures.} Unless specified otherwise, the dilation and stride are 1 for convolutional layers. The filter size is fixed to 5. The layers marked with $*$ are copied from the teacher to learners after the teacher training. id refers to the identity map. The TCN-N architecture uses dilated causal convolution~\cite{bai2018empirical}. Each convolutional layer is followed by a relu activation function.}  \label{tab:netarchitectures}
\end{table*}
\begin{table*}
\centering
 \begin{tabular}{|c|c|c|c|c|c|}
\hline 
 Model & seq. length & \# channels & \# param. & SGD time (s) \\ \hline
TCN-1  & 1 & 60 & 161960 & 9.22e-3 ($\pm$1.78e-3)\\ \hline
 TCN-20 & 20 & 44 & 158300 & 2.11e-2 ($\pm$1.24e-3)\\ \hline
 TCN-100 & 100 & 34 & 158070& 5.07e-2 ($\pm$1.94e-3)\\ \hline
 GRU & 100* & - & 159640 & 1.52e-1 ($\pm$1.89e-2) \\ \hline
\end{tabular}
\caption{\textbf{Network parameter settings and the training time for student policies.} SGD time refers to the computation time required for one stochastic gradient descent update with the batch size given in Table~S8. The computation times are presented as empirical means
with standard deviations. *The sequence length for the GRU network stands for the sequence length used for Truncated BPTT~\cite{williams1990efficient}.} \label{tab:netarchitectures2}
\end{table*}

\begin{table*}
\centering
 \begin{tabular}{|c|c|}
\hline 
 Parameter & Value  \\ \hline
 discount factor & 0.995 \\ \hline 
 KL-d threshold & 0.01  \\ \hline 
 max. episode length & 400 \\ \hline 
 CG damping & 1e-1 \\ \hline 
 CG iteration & 50 \\ \hline 
 discount factor & 0.995 \\ \hline 
batch size & 80000 \\ \hline 
total iterations& 10000 \\ \hline 
\end{tabular}
\caption{\textbf{Hyperparameters for teacher policy training.}}
\label{tab:hyperparametersteacher}
\end{table*}

\begin{table*}
\centering
 \begin{tabular}{|c|c|c|}
\hline 
 Parameter & TCN-N & GRU  \\ \hline
initial learning rate & 5e-4 & 2e-4 \\ \hline 
learning rate decay & \multicolumn{2}{|c|}{exp(0.995, 100)} \\ \hline
 max. episode length &\multicolumn{2}{|c|}{400} \\ \hline 
 batch size & 20000& 10000 \\ \hline 
 minibatches & \multicolumn{2}{|c|}{5} \\ \hline
 epochs &\multicolumn{2}{|c|}{4}\\ \hline 
 total iteration &\multicolumn{2}{|c|}{4000}\\ \hline 
\end{tabular}
\caption{\textbf{Hyperparameters for student policy training.} exp(a,b) denotes exponential decay, which is defined as $lr_0 * a^{updates/b}$. The Adam~\cite{kingma2014adam} optimizer is used.}
 \label{tab:hyperparameterslearner}
\end{table*}

\begin{table*}
\centering
 \begin{tabular}{|c|c|}
\hline 
 Parameter & values \\ \hline
initial learning rate & 1e-4 \\ \hline 
learning rate decay & exp(0.99, 100) \\ \hline
 batch size & 20000 \\ \hline 
 minibatches & 2\\ \hline
 epochs & 10\\ \hline 
 total iteration & 1000\\ \hline 
 weight decay & $l2$-norm, 1e-4 \\ \hline
\end{tabular}
\caption{\textbf{Hyperparameters for decoder training.} exp(a,b) denotes exponential decay, which is defined as $lr_0 * a^{updates/b}$. The Adam~\cite{kingma2014adam} optimizer is used.}
 \label{tab:hyperparametersdecoder}
\end{table*}


\begin{thebibliography}{10}

\bibitem{jenelten2019dynamic}
F.~Jenelten, J.~Hwangbo, F.~Tresoldi, C.~D. Bellicoso, M.~Hutter, Dynamic
  locomotion on slippery ground, {\it IEEE Robotics and Automation Letters\/}
  4170--4176 (2019).

\bibitem{bledt2018contact}
G.~Bledt, P.~M. Wensing, S.~Ingersoll, S.~Kim, Contact model fusion for
  event-based locomotion in unstructured terrains, {\it 2018 IEEE International
  Conference on Robotics and Automation (ICRA)\/} (IEEE, 2018).

\bibitem{focchi2020heuristic}
M.~Focchi, R.~Orsolino, M.~Camurri, V.~Barasuol, C.~Mastalli, D.~G. Caldwell,
  C.~Semini.
\newblock Heuristic planning for rough terrain locomotion in presence of
  external disturbances and variable perception quality.
\newblock {\it Advances in Robotics Research: From Lab to Market\/} (Springer,
  2020),  165--209.

\bibitem{Reher2019}
J.~Reher, W.~Ma, A.~D. Ames, Dynamic walking with compliance on a {Cassie}
  bipedal robot, {\it European Control Conference\/},  2589--2595 ({IEEE},
  2019).

\bibitem{gong2019feedback}
Y.~Gong, R.~Hartley, X.~Da, A.~Hereid, O.~Harib, J.~Huang, J.~W. Grizzle,
  Feedback control of a {Cassie} bipedal robot: Walking, standing, and riding a
  {Segway}, {\it American Control Conference\/},  4559--4566 (IEEE, 2019).

\bibitem{hwangbo2016probabilistic}
J.~Hwangbo, C.~D. Bellicoso, P.~Fankhauser, M.~Huttery, Probabilistic foot
  contact estimation by fusing information from dynamics and
  differential/forward kinematics, {\it 2016 IEEE/RSJ International Conference
  on Intelligent Robots and Systems (IROS)\/},  3872--3878 (IEEE, 2016).

\bibitem{camurri2017probabilistic}
M.~Camurri, M.~Fallon, S.~Bazeille, A.~Radulescu, V.~Barasuol, D.~G. Caldwell,
  C.~Semini, Probabilistic contact estimation and impact detection for state
  estimation of quadruped robots, {\it IEEE Robotics and Automation Letters\/}
  1023--1030 (2017).

\bibitem{focchi2018slip}
M.~Focchi, V.~Barasuol, M.~Frigerio, D.~G. Caldwell, C.~Semini.
\newblock Slip detection and recovery for quadruped robots.
\newblock {\it Robotics Research\/} (Springer, 2018),  185--199.

\bibitem{Bloesch2013}
M.~Bl{\"{o}}sch, C.~Gehring, P.~Fankhauser, M.~Hutter, M.~A. Hoepflinger,
  R.~Siegwart, State estimation for legged robots on unstable and slippery
  terrain, {\it 2013 {IEEE/RSJ} International Conference on Intelligent Robots
  and Systems\/},  6058--6064 ({IEEE}, 2013).

\bibitem{gehring2015dynamic}
C.~Gehring, C.~D. Bellicoso, S.~Coros, M.~Bloesch, P.~Fankhauser, M.~Hutter,
  R.~Siegwart, Dynamic trotting on slopes for quadrupedal robots, {\it 2015
  IEEE/RSJ International Conference on Intelligent Robots and Systems
  (IROS)\/},  5129--5135 (IEEE, 2015).

\bibitem{hartley2018legged}
R.~Hartley, J.~Mangelson, L.~Gan, M.~G. Jadidi, J.~M. Walls, R.~M. Eustice,
  J.~W. Grizzle, Legged robot state-estimation through combined forward
  kinematic and preintegrated contact factors, {\it 2018 IEEE International
  Conference on Robotics and Automation (ICRA)\/},  1--8 (IEEE, 2018).

\bibitem{hwangbo2019learning}
J.~Hwangbo, J.~Lee, A.~Dosovitskiy, D.~Bellicoso, V.~Tsounis, V.~Koltun,
  M.~Hutter, Learning agile and dynamic motor skills for legged robots, {\it
  Science Robotics\/} p. eaau5872 (2019).

\bibitem{haarnoja2018learning}
T.~Haarnoja, S.~Ha, A.~Zhou, J.~Tan, G.~Tucker, S.~Levine, Learning to walk via
  deep reinforcement learning, {\it Robotics: Science and Systems\/} (2019).

\bibitem{xie2019iterative}
Z.~Xie, P.~Clary, J.~Dao, P.~Morais, J.~Hurst, M.~van~de Panne, Learning
  locomotion skills for {Cassie}: Iterative design and sim-to-real, {\it
  Conference on Robot Learning\/} (2019).

\bibitem{lee2019robust}
J.~Lee, J.~Hwangbo, M.~Hutter, Robust recovery controller for a quadrupedal
  robot using deep reinforcement learning, {\it arXiv:1901.07517\/}  (2019).

\bibitem{tan2018sim}
J.~Tan, T.~Zhang, E.~Coumans, A.~Iscen, Y.~Bai, D.~Hafner, S.~Bohez,
  V.~Vanhoucke, Sim-to-real: Learning agile locomotion for quadruped robots,
  {\it Robotics: Science and Systems\/} (2018).

\bibitem{yang2019data}
Y.~Yang, K.~Caluwaerts, A.~Iscen, T.~Zhang, J.~Tan, V.~Sindhwani, Data
  efficient reinforcement learning for legged robots, {\it Conference on Robot
  Learning\/} (2019).

\bibitem{ha2020learning}
S.~Ha, P.~Xu, Z.~Tan, S.~Levine, J.~Tan, Learning to walk in the real world
  with minimal human effort, {\it arXiv:2002.08550\/}  (2020).

\bibitem{Peng2020}
X.~B. Peng, E.~Coumans, T.~Zhang, T.-W. Lee, J.~Tan, S.~Levine, Learning agile
  robotic locomotion skills by imitating animals, {\it arXiv:2004.00784\/}
  (2020).

\bibitem{hutter2016anymal}
M.~Hutter, C.~Gehring, D.~Jud, A.~Lauber, C.~D. Bellicoso, V.~Tsounis,
  J.~Hwangbo, K.~Bodie, P.~Fankhauser, M.~Bloesch, R.~Diethelm, S.~Bachmann,
  A.~Melzer, M.~A. H{\"{o}}pflinger, {ANYmal} - a highly mobile and dynamic
  quadrupedal robot, {\it {IEEE/RSJ} International Conference on Intelligent
  Robots and Systems\/},  38--44 ({IEEE}, 2016).

\bibitem{peng2018sim}
X.~B. Peng, M.~Andrychowicz, W.~Zaremba, P.~Abbeel, Sim-to-real transfer of
  robotic control with dynamics randomization, {\it IEEE International
  Conference on Robotics and Automation (ICRA)\/} (IEEE, 2018).

\bibitem{bai2018empirical}
S.~Bai, J.~Z. Kolter, V.~Koltun, An empirical evaluation of generic
  convolutional and recurrent networks for sequence modeling, {\it
  arXiv:1803.01271\/}  (2018).

\bibitem{chenlearning}
D.~Chen, B.~Zhou, V.~Koltun, P.~Kr\"ahenb\"uhl, Learning by cheating, {\it
  Conference on Robot Learning\/} (2019).

\bibitem{brant2017minimal}
J.~C. Brant, K.~O. Stanley, Minimal criterion coevolution: a new approach to
  open-ended search, {\it Genetic and Evolutionary Computation Conference\/},
  67--74 (2017).

\bibitem{wang2019paired}
R.~Wang, J.~Lehman, J.~Clune, K.~O. Stanley, Paired open-ended trailblazer
  (poet): Endlessly generating increasingly complex and diverse learning
  environments and their solutions, {\it arXiv:1901.01753\/}  (2019).

\bibitem{bellicoso2018dynamic}
C.~D. Bellicoso, F.~Jenelten, C.~Gehring, M.~Hutter, Dynamic locomotion through
  online nonlinear motion optimization for quadrupedal robots, {\it IEEE
  Robotics and Automation Letters\/}  2261--2268 (2018).

\bibitem{fankhauser2014robot}
P.~Fankhauser, M.~Bloesch, C.~Gehring, M.~Hutter, R.~Siegwart.
\newblock Robot-centric elevation mapping with uncertainty estimates.
\newblock {\it Mobile Service Robotics\/} (World Scientific, 2014),  433--440.

\bibitem{collins2005efficient}
S.~Collins, A.~Ruina, R.~Tedrake, M.~Wisse, Efficient bipedal robots based on
  passive-dynamic walkers, {\it Science\/}  1082--1085 (2005).

\bibitem{vision60blind}
{Ghost Robotics}, Vision 60: Latest blind-mode stress testing of {V60} legged
  robot, \url{www.youtube.com/watch?v=tQsLauQWp8M} (2019).

\bibitem{hwangbo2018per}
J.~Hwangbo, J.~Lee, M.~Hutter, Per-contact iteration method for solving contact
  dynamics, {\it IEEE Robotics and Automation Letters\/}  895--902 (2018).

\bibitem{coumans2013bullet}
E.~Coumans, others, Bullet physics library, Open source:
  \url{bulletphysics.org} (2013).

\bibitem{smith2005open}
R.~Smith, others, Open dynamics engine, Open source: \url{ode.org} (2005).

\bibitem{Alexander2003}
R.~M. Alexander, {\it Principles of Animal Locomotion\/} (Princeton University
  Press, 2003).

\bibitem{iscen2018policies}
A.~Iscen, K.~Caluwaerts, J.~Tan, T.~Zhang, E.~Coumans, V.~Sindhwani,
  V.~Vanhoucke, Policies modulating trajectory generators, {\it Conference on
  Robot Learning\/},  916--926 (2018).

\bibitem{barasuol2013reactive}
V.~Barasuol, J.~Buchli, C.~Semini, M.~Frigerio, E.~R. De~Pieri, D.~G. Caldwell,
  A reactive controller framework for quadrupedal locomotion on challenging
  terrain, {\it 2013 IEEE International Conference on Robotics and
  Automation\/},  2554--2561 (IEEE, 2013).

\bibitem{schulman2015trust}
J.~Schulman, S.~Levine, P.~Abbeel, M.~Jordan, P.~Moritz, Trust region policy
  optimization, {\it International Conference on Machine Learning\/},
  1889--1897 (2015).

\bibitem{bloesch2013state}
M.~Bloesch, M.~Hutter, M.~A. Hoepflinger, S.~Leutenegger, C.~Gehring, C.~D.
  Remy, R.~Siegwart, State estimation for legged robots-consistent fusion of
  leg kinematics and imu, {\it Robotics\/}  17--24 (2013).

\bibitem{ross2011reduction}
S.~Ross, G.~Gordon, D.~Bagnell, A reduction of imitation learning and
  structured prediction to no-regret online learning, {\it International
  Conference on Artificial Intelligence and Statistics\/},  627--635 (2011).

\bibitem{florensa2018automatic}
C.~Florensa, D.~Held, X.~Geng, P.~Abbeel, Automatic goal generation for
  reinforcement learning agents, {\it International Conference on Machine
  Learning\/},  1514--1523 (2018).

\bibitem{lehman2010revising}
J.~Lehman, K.~O. Stanley, Revising the evolutionary computation abstraction:
  minimal criteria novelty search, {\it Genetic and Evolutionary Computation
  Conference\/},  103--110 (2010).

\bibitem{matiisen2019teacher}
T.~Matiisen, A.~Oliver, T.~Cohen, J.~Schulman, Teacher-student curriculum
  learning, {\it IEEE transactions on neural networks and learning systems\/}
  (2019).

\bibitem{yu2018learning}
W.~Yu, G.~Turk, C.~K. Liu, Learning symmetric and low-energy locomotion, {\it
  ACM Transactions on Graphics (TOG)\/} p. 144 (2018).

\bibitem{akkaya2019solving}
I.~Akkaya, M.~Andrychowicz, M.~Chociej, M.~Litwin, B.~McGrew, A.~Petron,
  A.~Paino, M.~Plappert, G.~Powell, R.~Ribas, others, Solving rubik's cube with
  a robot hand, {\it arXiv:1910.07113\/}  (2019).

\bibitem{chavez2018learning}
R.~O. Chavez-Garcia, J.~Guzzi, L.~M. Gambardella, A.~Giusti, Learning ground
  traversability from simulations, {\it IEEE Robotics and Automation Letters\/}
   1695--1702 (2018).

\bibitem{kendall2017uncertainties}
A.~Kendall, Y.~Gal, What uncertainties do we need in bayesian deep learning for
  computer vision?, {\it Advances in neural information processing systems\/},
  5574--5584 (2017).

\bibitem{simonyan2013deep}
K.~Simonyan, A.~Vedaldi, A.~Zisserman, Deep inside convolutional networks:
  Visualising image classification models and saliency maps, {\it
  arXiv:1312.6034\/}  (2013).

\bibitem{pratt1995series}
G.~A. Pratt, M.~M. Williamson, Series elastic actuators, {\it IEEE/RSJ
  International Conference on Intelligent Robots and Systems\/},  399--406
  (1995).

\bibitem{smelik2009survey}
R.~M. Smelik, K.~J. De~Kraker, T.~Tutenel, R.~Bidarra, S.~A. Groenewegen, A
  survey of procedural methods for terrain modelling, {\it Proceedings of the
  CASA Workshop on 3D Advanced Media In Gaming And Simulation (3AMIGAS)\/},
  25--34 (2009).

\bibitem{lagae2010survey}
A.~Lagae, S.~Lefebvre, R.~Cook, T.~DeRose, G.~Drettakis, D.~S. Ebert, J.~P.
  Lewis, K.~Perlin, M.~Zwicker, A survey of procedural noise functions, {\it
  Computer Graphics Forum\/},  2579--2600 (Wiley Online Library, 2010).

\bibitem{chung2014empirical}
J.~Chung, C.~Gulcehre, K.~Cho, Y.~Bengio, Empirical evaluation of gated
  recurrent neural networks on sequence modeling, {\it arXiv:1412.3555\/}
  (2014).

\bibitem{williams1990efficient}
R.~J. Williams, J.~Peng, An efficient gradient-based algorithm for on-line
  training of recurrent network trajectories, {\it Neural Computation\/}
  490--501 (1990).

\bibitem{kingma2014adam}
D.~P. Kingma, J.~Ba, Adam: A method for stochastic optimization, {\it
  International Conference on Learning Representations\/} (2015).

\end{thebibliography}
\end{document}